\documentclass[letterpaper]{article} 
\usepackage{aaai25}  
\usepackage{times}  
\usepackage{helvet}  
\usepackage{courier}  
\usepackage[hyphens]{url}  
\usepackage{graphicx} 
\usepackage{natbib}  
\usepackage{caption} 
\usepackage{algorithm}
\usepackage{algorithmic}

\usepackage{newfloat}
\usepackage{listings}
\DeclareCaptionStyle{ruled}{labelfont=normalfont,labelsep=colon,strut=off} 
\floatstyle{ruled}
\newfloat{listing}{tb}{lst}{}
\floatname{listing}{Listing}
\title{Regional Expected Improvement for Efficient Trust Region Selection \\ 
in High-Dimensional Bayesian Optimization}
\author{
    Nobuo Namura\textsuperscript{\rm 1},
    Sho Takemori\textsuperscript{\rm 1}
}
\affiliations{
    \textsuperscript{\rm 1}Fujitsu Limited, Kawasaki, Japan\\
    \{namura.nobuo, takemori.sho\}@fujitsu.com
}

\usepackage{bibentry}

\usepackage{xcolor}

\usepackage{amssymb,amsmath,bm,amsthm}
\usepackage{cleveref}
\usepackage{dsfont}

\theoremstyle{plain}
\newtheorem{theorem}{Theorem}[section]
\newtheorem{proposition}[theorem]{Proposition}

\theoremstyle{definition}

\theoremstyle{remark}

\crefname{theorem}{Theorem}{Theorems}
\crefname{proposition}{Proposition}{Propositions}
\crefname{lemma}{Lemma}{Lemmas}
\crefname{definition}{Definition}{Definitions}
\crefname{remark}{Remark}{Remarks}

\DeclareMathOperator{\argmax}{argmax}
\DeclareMathOperator{\argmin}{argmin}

\newcommand{\RR}{\mathbb{R}}

\newcommand{\ex}[2][{}]{\mathbb{E}_{#1}\left[#2\right]}

\newcommand{\indc}{\mathds{1}}

\newcommand{\cA}{\mathcal{A}}

\newcommand{\cF}{\mathcal{F}}

\newcommand{\cH}{\mathcal{H}}

\newcommand{\cS}{\mathcal{S}}
\newcommand{\cT}{\mathcal{T}}

\newcommand{\extf}[1]{\widetilde{#1}}
\newcommand{\rbf}{F}
\newcommand{\totbgt}{N}

\begin{document}

\maketitle

\begin{abstract}
Real-world optimization problems often involve complex objective functions with costly evaluations. While Bayesian optimization (BO) with Gaussian processes is effective for these challenges, it suffers in high-dimensional spaces due to performance degradation from limited function evaluations. To overcome this, simplification techniques like dimensionality reduction have been employed, yet they often rely on assumptions about the problem characteristics, potentially underperforming when these assumptions do not hold. Trust-region-based methods, which avoid such assumptions, focus on local search but risk stagnation in local optima. In this study, we propose a novel acquisition function, regional expected improvement (REI), designed to enhance trust-region-based BO in medium to high-dimensional settings. REI identifies regions likely to contain the global optimum, improving performance without relying on specific problem characteristics. We provide a theoretical proof that REI effectively identifies optimal trust regions and empirically demonstrate that incorporating REI into trust-region-based BO outperforms conventional BO and other high-dimensional BO methods in medium to high-dimensional real-world problems.
\end{abstract}

%

\section{Introduction}
Real-world optimization problems often involve objective functions where gradients are difficult to obtain, and evaluations are time-consuming. Typical examples include the design optimization of industrial products like aircraft \cite{liem2015surrogate} and automobiles \cite{moustapha2015adaptive}, drug discovery \cite{shields2021bayesian}, materials development \cite{zhang2020bayesian}, and the tuning of hyperparameters in machine learning models \cite{perrone2021fair}. For such optimization problems, surrogate-based methods that can optimize with only tens to a few thousand of objective function evaluations are widely used. In particular, Bayesian optimization (BO) \cite{Mockus1978} using Gaussian processes (GPs) \cite{matheron1963principles} is frequently employed because it effectively balances the exploitation of the best-known solutions and the exploration of uncharted regions. This method constructs a GP model to represent the probability distribution of objective function values using a small number of evaluated sample points. It then iteratively generates new sample points by optimizing an acquisition function, which quantifies the utility of any point in the design space based on the GP model.

However, conventional BO using global GP models suffers from performance degradation as the number of design variables increases, making it primarily suitable for problems up to about 10 dimensions. For medium to high-dimensional black-box optimization problems exceeding 10 dimensions, the exponentially increasing design space poses a challenge due to the limited number of samples available for objective function evaluations \cite{binois2022survey}. As a result, problem simplification approaches are often applied. Common techniques include local exploration using GP models within trust regions \cite{regis2016trust}, dimensionality reduction assuming a sparse set of effective design variables \cite{wang2016bayesian}, reducing sensitivity in non-essential dimensions \cite{eriksson2021high}, simplifying GP models to capture only global trends under the assumption of a simple objective function \cite{hvarfner2024vanilla}, and solving subproblems under the assumption of separable design variables \cite{duvenaud2011additive}.

These methods have enabled improvements in objective function values even for problems with hundreds of dimensions. However, methods that rely on assumptions about problem characteristics may not perform well on problems that deviate from these assumptions. In practice, the characteristics of a problem are often unknown until it is solved, which raises concerns about the generalization performance of these methods. Even for problems with tens of dimensions, they underperform compared to standard BO with global GP models \cite{li2017high}. 
High-dimensional BO methods that do not rely on such assumptions are limited \cite{oh2018bock, gupta2020trading}. Among these, methods that use trust regions have been gaining attention. However, these methods primarily focus on local search, which raises concerns about stagnation in local optima. For instance, trust region BO (TuRBO) \cite{eriksson2019scalable} explores multiple randomly defined trust regions simultaneously to capture the global optimum, but efficiently obtaining the global optimum remains challenging due to the vast design space in high-dimensional problems.

In this study, we propose a novel acquisition function, regional expected improvement (REI), to enhance trust-region-based BO for global optimization in high-dimensional problems. By introducing REI into trust-region-based BO, we can efficiently solve medium to high-dimensional problems without making assumptions about the characteristics of the problem. The specific contributions of this study are as follows:

\begin{itemize}
\item We propose REI, an acquisition function designed to identify regions likely to include the global optimum and to escape from local optima.
\item We theoretically prove that REI effectively identifies an optimal trust region.
\item We empirically show that incorporating REI into TuRBO outperforms conventional BO and various high-dimensional BO methods, without relying on assumptions about problem characteristics.
\end{itemize}

\noindent The code for REI is available at: \url{https://github.com/nobuo-namura/regional-expected-improvement}. Appendices for this paper can be found in \citet{namura2024rei}.

\section{Related Works}
In high-dimensional BO, the sample density decreases across the extensive design space, necessitating approaches that reduce the complexity of GP models.

A common strategy assumes that in high-dimensional design spaces, some dimensions contribute minimally to the variation in the objective function. This strategy compresses the design space into an active subspace comprising only the effective dimensions. By constructing a GP model on the compressed design space, model complexity is reduced.
The random embedding BO (REMBO) \cite{wang2016bayesian} and hashing-enhanced subspace BO (HeSBO) \cite{nayebi2019framework} use random matrices for dimensionality reduction and demonstrate theoretical utility when effective dimensions are limited. Alternatively, the sparse axis-aligned subspace BO (SAASBO) \cite{eriksson2021high} simplifies the model by using a sparse prior distribution that reduces the sensitivity of the kernel function to ineffective dimensions by decreasing their inverted length scale. The vanilla BO with $D$-scaled $p(\ell)$ (DSP) \cite{hvarfner2024vanilla} assumes that even in high-dimensional problems, the landscape of the objective function is simple. By setting a prior distribution that increases the length scale with the dimension size and simplifying the variance parameters of a GP model, it prevents model complexity from escalating. This approach achieves a simplification similar to SAASBO but with reduced computational cost.

Another notable approach involves using local GP models instead of a global GP model. This approach constructs GP models using sample points distributed within full-dimensional subregions of the design space or restricts sample addition to specific subregions. This local modeling approach suppresses complexity increases and accurately captures the local shape of the objective function. TuRBO employs a trust region centered around the best existing sample point, progressively reducing the size of the trust region when no improvement in the objective function is observed. This method captures detailed local structures incrementally while advancing the optimization process. When the trust region becomes too small, the method restarts with a new trust region centered around the best randomly generated sample point. An extension of TuRBO, BO with adaptively expanding subspaces (BAxUS) \cite{papenmeier2022increasing}, combines dimensionality reduction with TuRBO's search, gradually increasing the dimensions used for model construction instead of restarting the trust region. Additionally, CASMOPOLITAN \cite{wan2021think} extends TuRBO for discrete optimization and mixed-integer problems. Similarly, Bounce \cite{papenmeier2023bounce} extends BAxUS to handle discrete variables and incorporates strategies to reduce the number of samples needed for optimization in lower-dimensional spaces.

\section{Preliminaries}
We consider obtaining the global optimum of the minimization problem for the objective function $f$ as shown below:
\begin{equation}
    \label{eq:min-prob}
    \mathbf{x}^{*} = \argmin_{\mathbf{x} \in \Omega} f(\mathbf{x}),
\end{equation}
where $\Omega = [0,1]^D$ represents a $D$-dimensional design space. This study focuses on problems with $D>10$, where the performance of standard GP-based BO is known to deteriorate. For maximization problems of an objective function $f'$, we consider solving the corresponding minimization problem by setting $f=-f'$. In this section, we discuss the foundational concepts of GP, BO, and TuRBO, which are prerequisites for introducing REI.

\subsection{Gaussian Process}
For a mean $m: \Omega \rightarrow \RR$ and a covariance (kernel) function $k : \Omega \times \Omega  \rightarrow \RR$,
a GP $\hat{f} \sim \mathcal{GP}(m(\mathbf{x}), k(\mathbf{x}, \mathbf{x}'))$ is a random function (or stochastic process) characterized by the following properties:
(i) for any sequence $\mathbf{x}_1, \dots, \mathbf{x}_i \in \Omega$ with $i \ge 1$, $\hat{f}(\mathbf{x}_1), \dots, \hat{f}(\mathbf{x}_i)$ are jointly Gaussian,
(ii) for any $\mathbf{x}, \mathbf{x}' \in \Omega$, the mean $\ex{\hat{f}(\mathbf{x})}$ and covariance $\mathrm{Cov}[\hat{f}(\mathbf{x}), \hat{f}(\mathbf{x}')]$ are given as 
$m(\mathbf{x})$ and $ k(\mathbf{x}, \mathbf{x}')$, respectively.
By providing sample points $\mathcal{D}_t = \{ \mathbf{x}_{i}, f_{i}\}_{i=1}^{t}$ consisting of pairs of specific design variables $\mathbf{x}_{i}$ and objective function values $f_{i}$ to this model, we can derive the functions representing the posterior mean $\hat{f}(\mathbf{x}|\mathcal{D}_t)$ and variance $\hat{\sigma}^{2}(\mathbf{x}|\mathcal{D}_t)$ as a surrogate of the objective function. In this study, assuming the mean function is zero, it is necessary to standardize the objective function values of the sample points before generating the GP model. Additionally, we assume that the observed values of the function include noise following $\mathcal{N}(0,\sigma_{n}^{2})$. The noise variance $\sigma_{n}^{2}$ and the kernel hyperparameters are determined by maximum a posteriori (MAP) estimation.

\subsection{Bayesian Optimization}
In BO, the process iteratively approximates the objective function using the GP model based on existing sample points and adds new sample points by optimizing the acquisition function. The acquisition function is designed to balance exploitation, which involves adding new sample points in regions likely to contain the optimal solution to improve the objective function, and exploration, which involves adding new sample points in regions with low sample density to improve the approximation accuracy of the GP model.

Prominent acquisition functions include expected improvement (EI) \cite{Mockus1978} and Thompson sampling (TS) \cite{thompson1933likelihood}. The EI at a location $\mathbf{x}$ on the GP model, assuming a normal distribution, represents the expected value of the objective function improvement over the best-known objective function value $f_{\text{ref}} \in \mathcal{D}_t$, and is calculated as follows. 
\begin{align*}
    \label{eq:ei}
    \text{EI}(\mathbf{x}) &= \int_{-\infty}^{f_{\text{ref}}} (f_{\text{ref}} - f)\phi(f) df \\
    &= (f_{\text{ref}} - \hat{f}(\mathbf{x}|\mathcal{D}_t)) \Phi(f_{\text{ref}}) + \hat{\sigma}(\mathbf{x}|\mathcal{D}_t) \phi(f_{\text{ref}})
\end{align*}
where $\phi$ and $\Phi$ denote the probability density function and cumulative distribution function, respectively, of the normal distribution following $\mathcal{N}(\hat{f}(\mathbf{x}|\mathcal{D}_t),\hat{\sigma}^{2}(\mathbf{x}|\mathcal{D}_t))$. EI is often maximized using gradient methods, but the gradient can vanish in regions with small improvement. Therefore, LogEI \cite{ament2023unexpected}, which uses the logarithm of the improvement, has been proposed. Additionally, a Monte Carlo acquisition function qEI \cite{wang2020parallel, balandat2020botorch}, which calculates EI via Monte Carlo sampling, is utilized to add multiple sample points simultaneously.

In contrast, TS involves randomly sampling functions $\text{TS} \sim \mathcal{GP}(0, k(\mathbf{x}, \mathbf{x}')|\mathcal{D}_t)$ from the posterior distribution of the GP and then adding new samples by minimizing the sampled function in the case of minimization problems. Although the latter part is greedy, the functions sampled from the posterior distribution are more likely to have low objective function values in regions where the GP uncertainty $\hat{\sigma}^{2}(\mathbf{x})$ is high, thus balancing exploration and exploitation.

\subsection{Trust Region Bayesian Optimization}
TuRBO is a representative high-dimensional Bayesian optimization method that uses trust regions. It is denoted as TuRBO-$m$ where $m$ represents the number of trust regions used in parallel. For simplicity, Algorithm \ref{alg:turbo} illustrates the procedure for TuRBO-1, where there is only one trust region and a single sample point is added in each iteration. Note that while the original TuRBO-1 uses TS as the acquisition function, this study also implements LogEI.

In TuRBO, $n$ initial sample points are generated uniformly at random across the entire design space, and a hypercube with side length $l=l_0$ centered around the sample point $\mathbf{c}$ with the minimum objective function value is defined as the trust region. A GP model is then constructed using the initial sample points, and the next sample point is selected by minimizing the TS acquisition function or maximizing the LogEI acquisition function within the trust region.

To update the trust region, the objective function value of the added sample point is compared with that of the existing sample points for the current trust region. If the minimum value of the objective function does not improve for $\tau_{\text{fail}}$ consecutive iterations, $l$ is halved, reducing the search area. Conversely, if the minimum value of the objective function improves for $\tau_{\text{succ}}$ consecutive iterations, $l$ is doubled, expanding the search area. If the minimum objective function value is updated by the added sample point, the center of the trust region is also updated to the location of that sample point. The actual side length of the trust region is adjusted for each dimension according to the length scale of the GP model. 

If $l$ becomes smaller than a predetermined minimum length $l_{min}$, the current trust region's search is terminated. The optimization is then restarted in a new trust region by generating initial sample points randomly, without using any information from previous sample points.

\begin{algorithm}[tb]
\caption{TuRBO-1}
\label{alg:turbo}
\textbf{Input}: Objective function $f$, Design space $\Omega$, Initial budget $n$, Total budget $N$ \\
\textbf{Parameter}: Initial trust region length $l_0$, Minimum trust region length $l_{min}$ \\
\textbf{Output}: Samples $\mathcal{D}_N = \{ \mathbf{x}_{i}, f_{i}\}_{i=1}^{N}$ \\
\begin{algorithmic}[1] 
\STATE $\mathcal{D}_0 = \varnothing$, $l = l_0$, $t = 0$
\WHILE{$t < N$}
    \IF{$t=0$ or $l < l_{min}$}
        \STATE Generate initial samples $\mathcal{D}^{TR} = \{ \mathbf{x}_{i}, f_{i}\}_{i=1}^{n}$
        \STATE Initialize trust region $\mathcal{TR}$ with length $l=l_0$ centered around $\mathbf{c}=\argmin_{\mathbf{x} \in \mathcal{D}^{TR}} f(\mathbf{x})$
        \STATE Augment $\mathcal{D}_{t+n} \leftarrow \mathcal{D}_{t} \cup \mathcal{D}^{TR}$
        \STATE $t \leftarrow t + n$
    \ELSE
        \STATE Generate a GP model $\hat{f}(\mathbf{x}|\mathcal{D}^{TR})$, $\hat{\sigma}^{2}(\mathbf{x}|\mathcal{D}^{TR})$
        \STATE Select a next sample using an acquisition function
        $\mathbf{x}_{t+1} = \argmin_{\mathbf{x} \in \mathcal{TR}} \text{TS}(\mathbf{x})$ \\
        \quad or \\
        $\mathbf{x}_{t+1} = \argmax_{\mathbf{x} \in \mathcal{TR}} \text{LogEI}(\mathbf{x})$ 
        \STATE Evaluate objective function $f_{t+1} = f(\mathbf{x}_{t+1})$
        \STATE Update $\mathbf{c}$ and $l$ of $\mathcal{TR}$
        \STATE Augment $\mathcal{D}^{TR} \leftarrow \mathcal{D}^{TR} \cup \{(\mathbf{x}_{t+1}, f_{t+1})\}$
        \STATE Augment $\mathcal{D}_{t+1} \leftarrow \mathcal{D}_{t} \cup \mathcal{D}^{TR}$
        \STATE $t \leftarrow t + 1$
    \ENDIF
\ENDWHILE
\STATE \textbf{return} $\mathcal{D}_{t}$
\end{algorithmic}
\end{algorithm}

\section{Proposed Method}
While TuRBO demonstrates excellent performance for high-dimensional optimization problems without requiring special assumptions, it predominantly focuses on local search and has limited strategies for obtaining the global optimum. In this section, we introduce REI, an approach specifically designed to efficiently discover trust regions that are likely to contain the global optimum. We describe the method for selecting the trust region when applying REI to TuRBO. It is important to note that REI is a general approach that can be applied to any optimization method utilizing trust regions, not just TuRBO.

\subsection{Region-Averaged Acquisition Functions}
Since TuRBO evaluates solutions hundreds of times within a single trust region once it is selected, the trust region should be determined based on macro features within the region, rather than the local landscape details of the GP. Therefore, instead of using conventional acquisition functions such as EI or upper confidence bound (UCB), which strongly depend on the local structure of the GP, a specialized acquisition function that aggregates the GP structure within the region is needed. 

This acquisition function requires converting the GP distribution within the trust region into a scalar value. We propose simply averaging commonly used acquisition functions within the trust region to achieve this scalarization.
Given the center of the trust region as $x_d$ and its length as $l_d$ for the $d$-th dimension, the trust region around the center $\mathbf{x}$ is defined as follows:
\begin{equation}
    \label{eq:tr}
    \mathcal{TR}_{\mathbf{x}} = \prod_{d=1}^D [\max(x_d - l_d/2, 0), \min(x_d + l_d/2, 1)].
\end{equation}
The region-averaged acquisition function is then defined as the mean value of the acquisition function within this trust region. 
While averaging the objective function is common in robust optimization \cite{ribaud2020robust}, this study instead uses the average of the acquisition function, which captures the uncertainty of the GP model rather than the objective function itself.

\subsection{Regional Expected Improvement}
A practical region-averaged acquisition function, REI, is defined as the mean value of EI within the trust region. However, the integral of EI within the trust region is analytically challenging to obtain, so this study employs Monte Carlo sampling as a substitute:
\begin{equation}
\label{eq:rei}
    \text{REI}(\mathbf{x}) = \frac{1}{V_{\mathcal{TR}_{\mathbf{x}}}} \int_{\mathcal{TR}_{\mathbf{x}}} \text{EI}(\hat{\mathbf{x}}) d\hat{\mathbf{x}} \approx \frac{1}{N_{x}} \sum_{j=1}^{N_{x}} \text{EI}(\hat{\mathbf{x}}_j),
\end{equation}
where $\hat{\mathbf{x}}_j \in \mathcal{TR}_{\mathbf{x}}$ is a Monte Carlo sample in the trust region. 
We primarily use TuRBO-1; however, to prepare for selecting multiple trust regions in TuRBO-$m$, we employ the Monte Carlo acquisition function variant qREI by replacing EI with qEI:
\begin{equation}
\label{eq:qrei}
    \text{qREI}(\mathbf{x}) = \frac{1}{N_x N_f} \sum_{j=1}^{N_x} \sum_{k=1}^{N_f} \max(f_{\text{ref}} - \text{TS}_k(\hat{\mathbf{x}}_j), 0),
\end{equation}
where $\text{TS}_k$ is the $k$-th randomly sampled function from the posterior distribution
of the GP.

\begin{figure}[t]
    \centering
    \includegraphics[width=0.8\columnwidth]{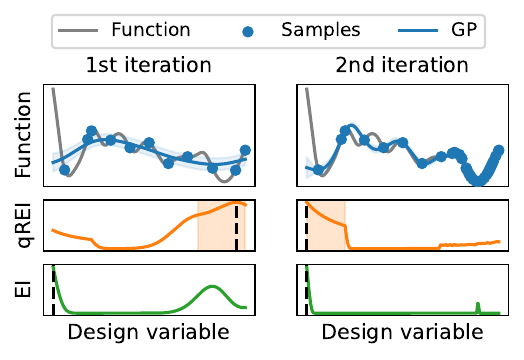}
    \caption{Comparison of qREI and EI distributions}
    \label{fig:concept}
\end{figure}

By using qREI, regions with greater potential for yielding good solutions can be identified. The left side of Fig. \ref{fig:concept} shows the results of calculating qREI and EI after uniformly generating initial samples for a one-dimensional minimization problem. EI exhibits a sharp peak on the left end and a broader peak on the right end. When using EI to select the trust region, aiming to locate points with the largest objective function improvement, the left end, where EI is maximized, is selected. 
In contrast, qREI averages EI within a defined range centered on each point (the orange-shaded region centered at the dashed line) to search for spaces with robust improvement. Consequently, qREI achieves its maximum at the right end, where EI has a broader peak. 
Once the local search in the trust region near the right end is completed, a second region selection occurs (right side of Fig. \ref{fig:concept}), and local search is conducted in the left-end region. 

In actual problems, EI is likely to have many local maxima, and qREI selects regions not only based on the width of each EI peak but also on the number of local optima within the region. Region selection using qREI initially prioritizes areas with broad valleys, ensuring good solutions even within local optima. As the search progresses, qREI shifts to selecting regions with deep valleys that have the potential to update known local optima.

\begin{algorithm}[t]
\caption{Trust region selection}
\label{alg:selection}
\textbf{Input}: Objective function $f$, Design space $\Omega$, Initial budget $n$, Samples $\mathcal{D}_t$ \\
\textbf{Parameter}: Initial trust region length $l_0$ \\
\textbf{Output}: Samples $\mathcal{D}^{TR}$ \\
\begin{algorithmic}[1] 
\IF{$\mathcal{D}_t = \varnothing$}
    \STATE Generate initial samples at random \\
    $\mathcal{D}_t = \mathcal{D}^{TR} = \{ \mathbf{x}_{i}, f_{i}\}_{i=1}^{n}$
    \STATE $t=n$
\ELSE
    \STATE $\mathcal{D}^{TR} = \varnothing$
\ENDIF
\STATE Generate a GP model $\hat{f}(\mathbf{x}|\mathcal{D}_t)$, $\hat{\sigma}^{2}(\mathbf{x}|\mathcal{D}_t)$
\STATE Select a trust region $\mathcal{TR}_{\mathbf{c}}$ with length $l=l_0$ using qREI \\
$\mathbf{c} = \argmax_{\mathbf{x} \in \Omega} \text{qREI}(\mathbf{x})$ \\
\STATE Generate samples in the selected trust region at random\\
$\{\mathbf{x}_i\}_{i=t+1}^{t+n-1} \in \mathcal{TR}_{\mathbf{c}}$
\STATE Evaluate the objective function and augment a dataset \\
$\mathcal{D}^{TR} \leftarrow \mathcal{D}^{TR} \cup \{\mathbf{c}, f(\mathbf{c})\} \cup \{\mathbf{x}_{i}, f(\mathbf{x}_{i})\}_{i=t+1}^{t+n-1}$
\STATE \textbf{return} $\mathcal{D}^{TR}$
\end{algorithmic}
\end{algorithm}

\subsection{Trust Region Selection}
The trust region selection using qREI is performed during the initialization and restart phases of TuRBO. Specifically, this is achieved by replacing the initial sample generation in line 4 of Algorithm \ref{alg:turbo} with the procedure outlined in Algorithm \ref{alg:selection}.
During these phases, a global GP model is generated using all existing sample points. The next trust regions are determined by fixing the trust region length to the initial value $l=l_0$ and varying the centers $\mathbf{x} \in \Omega$ to maximize qREI. 
Once the trust region is selected, TuRBO proceeds as described in Algorithm \ref{alg:turbo} using initial samples that include one point at the center of the selected trust region and $(n-1)$ points randomly generated within the trust region.

\section{Theoretical Analysis}
In the last section, we have defined region-averaged acquisition functions (specifically the qREI acquisition function) for identifying a trust region likely to include the global optimum.
In this section, we prove that these region-averaged acquisition functions enable effective exploration of an optimal trust region (i.e., region likely including the global optimum).
We provide proofs in Appendix \ref{apx:app-proof}.

\paragraph{Notation and Assumptions.}
Let $k: \RR^D \times \RR^D \rightarrow \RR$ be a positive definite kernel of the form $k(\mathbf{x}, \mathbf{x}') = \rbf(\mathbf{x} - \mathbf{x}')$ for any $\mathbf{x}, \mathbf{x}' \in \RR^D$, where 
$\rbf$ is a continuous function on $\RR^D$ with $\rbf \in L^1(\RR^D)$ and $L^1(\RR^D)$ is the space of $L^1$ integrable functions.
This represents a wide class of kernels including Mat\'ern, Gaussian, and rational quadratic kernels \cite[Chapter 6]{wendland2004scattered}.
We denote the associated reproducing kernel Hilbert space (RKHS) on $\Omega$ by $\cH_\rbf(\Omega)$ and
the RKHS norm of $f \in \cH_\rbf(\Omega)$ by $\|f\|_{\cH_\rbf(\Omega)}$.

\begin{figure}[t]
    \centering
    \includegraphics[width=0.6\columnwidth]{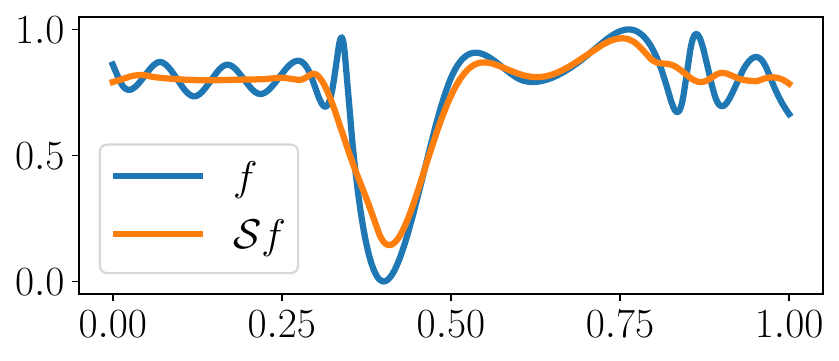}
    \caption{A simple example of the objective function $f$ and its region-averaged function $\cS f$.}
    \label{fig:obj_fn_and_region_fn}
\end{figure}

\paragraph{Region-Averaging Operator.}
To discuss an optimization problem regarding the region-averaged acquisition functions, 
we define an operator $\cS$ that transforms the objective function to a region-averaged objective function as follows.
For $f \in \cH_\rbf(\Omega)$, it is known that there exists a natural extension $\extf{f} : \RR^D \rightarrow \RR$ 
of $f$ (i.e., $\extf{f}(\mathbf{x}) = f(\mathbf{x})$ for any $\mathbf{x} \in \Omega$) with 
$\extf{f} \in \cH_\rbf(\RR^D)$
(we refer to Theorem 10.46 in \citet{wendland2004scattered}. See also \Cref{prop:rkhs-norm-restr}). 
For $f \in \cH_\rbf(\Omega)$, we define a region-averaged objective function $\cS f: \Omega \rightarrow \RR$ as 
\begin{equation*}
    \cS f (\mathbf{x}) :=
    \ex[\mathbf{u} \sim  \mathrm{Unif}(\mathcal{TR}_0)]{\extf{f}(\mathbf{x} + \mathbf{u})}, \quad \mathbf{x} \in \Omega.
\end{equation*}
Here $\mathcal{TR}_0$ is a trust region centered at $0 \in \RR^{D}$ defined as $\left(\prod_{i=1}^{D}[-l_i/2, l_i/2]\right)$
and $\mathrm{Unif}(\mathcal{TR}_0)$ denotes the uniform distribution on it.
In Fig. \ref{fig:obj_fn_and_region_fn}, we illustrate a simple example of the objective function $f$ 
and $\cS f$ in the case when $D=1$.
If $a: \RR^{D} \rightarrow \RR$ is a lower confidence bound (LCB) of the (extended) objective function $\extf{f}$,
then by definition, the region-averaged acquisition function $\cS a$ is a LCB of $\cS f$.
More generally, for any acquisition function $a$, we regard the region-averaged acquisition function $\cS a$ as an acquisition function for 
$\cS f$.


\paragraph{Problem Complexity of BO.}
For a Bayesian optimization algorithm $\cA$ and an objective function $f \in \cH_{\rbf}(\Omega)$, 
we define (cumulative) regret $R_\totbgt(f, \cA)$ by $\sum_{t=1}^\totbgt (f(\mathbf{x}_t) - f(\mathbf{x}^{*}))$, 
where $\mathbf{x}_t \in \Omega$ is a point selected by the algorithm $\cA$.
Following \cite{gupta2022regret,chowdhury2017kernelized}, 
we define 
\begin{equation}
    \label{eq:u_n-def}
    U_\totbgt(f) = \beta_{\totbgt}(f) \sqrt{\totbgt \gamma_\totbgt},
\end{equation}
where $\beta_{\totbgt}(f) := \sigma_n \sqrt{2(\gamma_{\totbgt-1} + 1 + \log (1/\delta))} + \| f\|_{\cH_{\rbf}(\Omega)}$, $\delta \in (0, 1)$, and 
$\gamma_\totbgt$ is the maximum information gain \citep{chowdhury2017kernelized}.
Then, it is known that popular algorithms such as GP-EI and GP-UCB (or IGP-UCB) have 
the following regret upper bound \cite{gupta2022regret,chowdhury2017kernelized, srinivas2010gaussian}:
That is, the following inequality holds with a probability at least $1 - \delta$:
\begin{align*}
    R_\totbgt(f, \cA) = O(U_\totbgt(f)),
\end{align*}
for $\cA = \text{GP-EI}$ and $ \text{IGP-UCB}$,
where $\delta \in (0, 1)$ is a parameter of the algorithms.
Therefore, we can regard $U_\totbgt(f)$ as the problem complexity of the Bayesian optimization problem for the objective function $f$.
\paragraph{Main Theorem.} 
The example illustrated in Fig. \ref{fig:obj_fn_and_region_fn} indicates that 
the complexity of the objective function $f$ is reduced if we take the operator $\cS$.
The following theorem formalizes this intuition for a wide class of kernels.
\begin{theorem}
    \label{thm:reduced-upper-bound}
    For any objective function $f \in \cH_\rbf(\Omega)$, let $U_\totbgt(f)$ be the regret upper bound defined as \eqref{eq:u_n-def}. Then, for any $\totbgt \ge 1$,
    $U_\totbgt(f)$ is reduced if we take the operator $\cS$, i.e., we have $\cS f \in \cH_F(\Omega)$ and the following inequality holds:
    \begin{equation*}
        U_\totbgt(\cS f)  \le U_\totbgt(f).
    \end{equation*}
\end{theorem}
\cref{thm:reduced-upper-bound} indicates that with region-averaged acquisition functions such as qREI, 
one can explore a globally optimal trust region more effectively compared 
to the original optimization problem regarding $f$ and shows a potential superiority over standard methods using global GP models.

\citet{wan2021think} also extended TuRBO and provided cumulative regret analysis (Theorems 3.3, 3.4, \citet{wan2021think}).
However, their regret analysis followed the standard proof techniques developed by \cite{srinivas2010gaussian},
and they have not provided a theoretical comparison to standard methods using global GP models such as GP-UCB \cite{srinivas2010gaussian}.

\section{Experiment}

\begin{figure*}[t]
\centering
\includegraphics[width=0.9\textwidth]{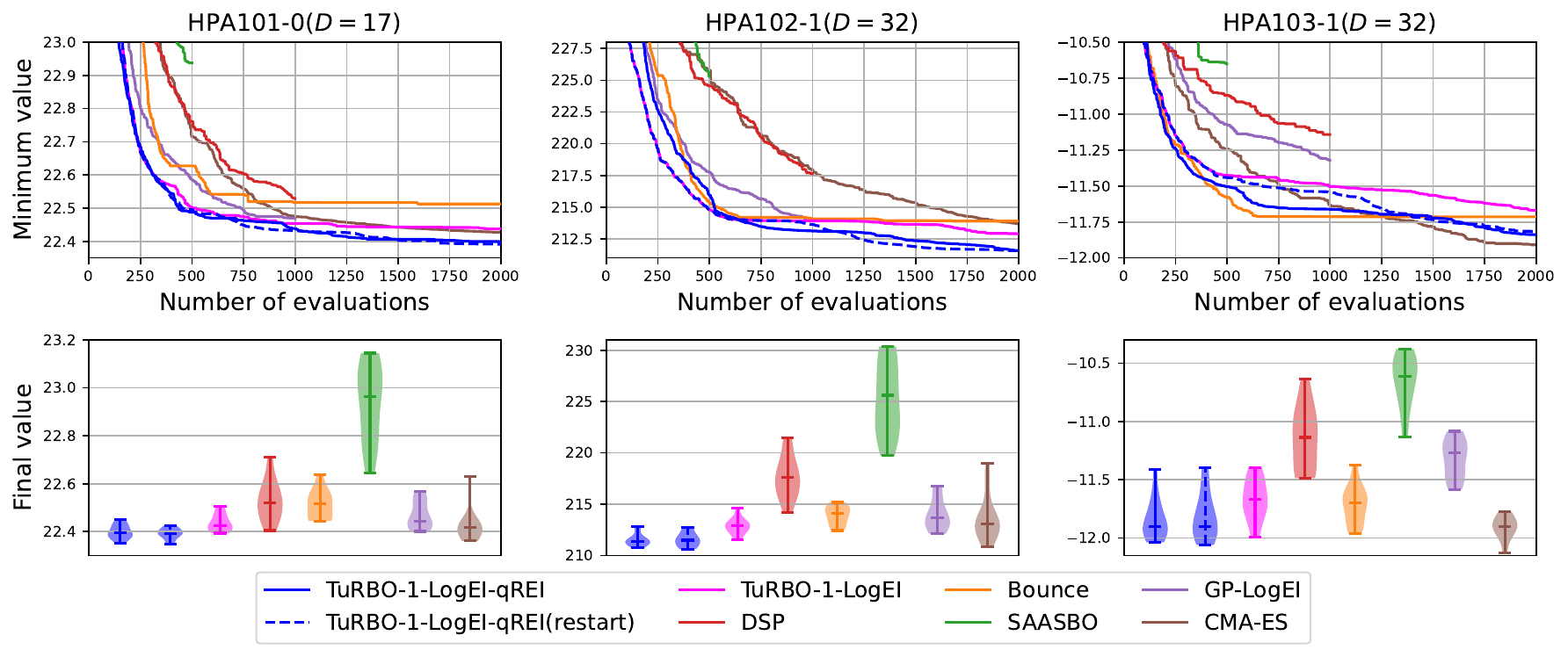} 
\caption{Histories and violin plots of objective function values for the medium-dimensional problems.}
\label{fig:mid-prob}
\end{figure*}

\subsection{Implementation Details}
We evaluate the effectiveness of qREI using TuRBO-1 implemented in Botorch \cite{balandat2020botorch} with LogEI as the acquisition function for local search (TuRBO-1-LogEI). 
Results for TuRBO-$m$ and TuRBO-1 with TS as the acquisition function are provided in Appendices \ref{apx:trubom} and \ref{apx:trubo-ts}.
For simplicity, qREI is always used for trust region selection with $N_x=128$ and $N_f=256$ used for Monte Carlo sampling. 
We use two versions of TuRBO with qREI: one where qREI is implemented as in Algorithm 2, used for both initialization and restarts, and another where standard TuRBO initialization is used, with qREI applied only during restarts. The former is denoted as TuRBO-1-LogEI-qREI, while the latter is referred to as TuRBO-1-LogEI-qREI(restart).

The parameters for TuRBO follow those of Botorch implementation with $\tau_{\text{succ}}=10$. Other key parameters, including $l_0=0.8$, $l_{min}=0.5^7$, and $\tau_{\text{fail}}=D$, are consistent with those in the original paper. 
Details of the other parameter settings are provided in Appendix \ref{apx:setting}.
\subsection{Baselines}
The baselines for comparison include TuRBO-1-LogEI (without qREI), along with DSP, Bounce, SAASBO, GP-LogEI, and CMA-ES \cite{hansen2001completely}. DSP and Bounce were implemented using the authors' original repositories. 
SAASBO and GP-LogEI adopted the Botorch implementation, with SAASBO using LogEI instead of EI as the acquisition function, and GP-LogEI using a global GP and LogEI for standard BO. For CMA-ES, we used the pycma \cite{hansen2019pycma} implementation through pymoo \cite{pymoo}.

\subsection{Benchmarks}
To address high-dimensional optimization problems where assumptions about problem characteristics do not hold, we use benchmark problems based on real-world applications. 
In evaluating the effectiveness of qREI, the number of restarts is considered crucial. To confirm the impact of qREI within a limited number of function evaluations, we mainly focus on design optimization problems for vehicles where noise influence is minimal. Specifically, we use human-powered aircraft (HPA) \cite{namura2024hpa}, rover trajectory planning \cite{wang2018batched}, and MOPTA08 problems. 
The HPA test suite includes nine single-objective problems. We used HPA101-0 ($D=17$), 102-1 ($D=32$), and 103-1 ($D=32$) as medium dimensional problems and 101-2 ($D=108$) as a high-dimensional problem. 
The rover trajectory planning problem ($D=60$) was used with a fixed noise term to ensure reproducibility, in order to clearly assess the impact of qREI. The MOPTA08 problem ($D=124$) was used in its unconstrained form, consistent with the previous study \cite{eriksson2021high}. These two problems were used as high-dimensional problems.

\subsection{Setup}
For all problems and methods except CMA-ES, the number of initial samples is $n=30$, and the batch size is 1. The population sizes of CMA-ES are determined by its own algorithm, implemented with the default settings. Due to computational resource constraints, the maximum number of function evaluations varies by method: $N=2000$ for the proposed methods, TuRBO, Bounce, and CMA-ES; $N=1000$ for DSP and GP-LogEI; and $N=500$ for SAASBO. The number of independent runs with different random seeds is 11 for all problems.
To compare the optimization results, we use the mean histories of the best objective function values with respect to the number of evaluations and violin plots of the final results. These visualizations allow us to evaluate the best method based on the number of evaluations and assess the variability in the performance of each method.

\begin{figure*}[t]
\centering
\includegraphics[width=0.9\textwidth]{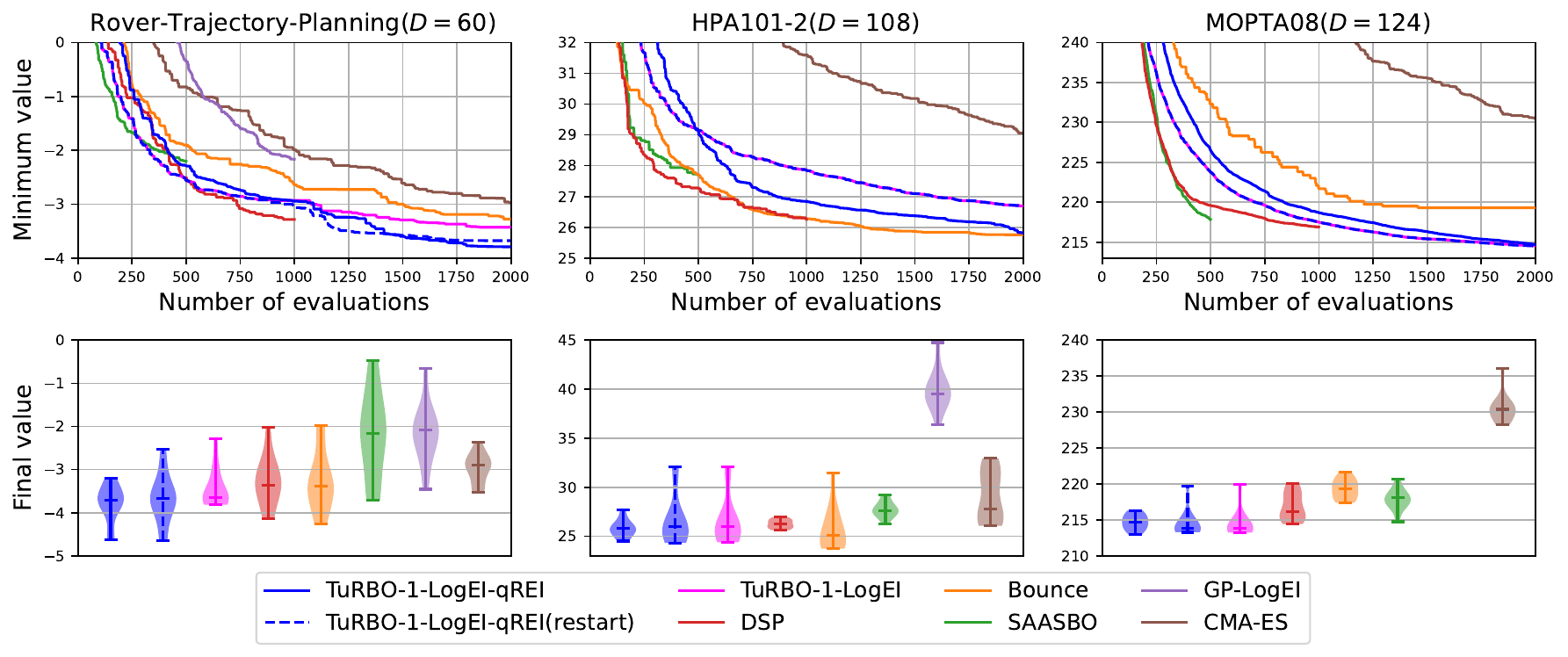} 
\caption{Histories and violin plots of objective function values for the high-dimensional problems.}
\label{fig:high-prob}
\end{figure*}

\subsection{Results}

\paragraph{Medium-Dimensional Problems.} 
First, we evaluate the effectiveness of qREI in mid-dimensional problems, where TuRBO excels and trust regions are restarted many times. 
As illustrated in Fig. \ref{fig:mid-prob}, the use of qREI enhances the performance of TuRBO-1-LogEI across all problems. Both variants of TuRBO-1-LogEI-qREI exhibit the best performance in HPA101-0 and HPA102-1. In these two problems, many methods underperform relative to GP-EI. However, TuRBO-1-LogEI-qREI, which significantly surpasses both TuRBO-1-LogEI and GP-EI (see Appendix \ref{apx:testing}), emerges as a practical approach for medium-dimensional problems.
In HPA103-1, unlike DSP, SAASBO, and GP-EI, which show significantly reduced performance, TuRBO-1-LogEI-qREI demonstrates excellent performance on par with Bounce and CMA-ES. It consistently exhibits high performance across different problems.
In Appendix \ref{apx:rei-vs-ei}, we experimentally demonstrate that region-averaged acquisition functions, including qREI, yield better solutions than those obtained through trust region selection based on conventional acquisition functions, such as EI or UCB.

\paragraph{High-Dimensional Problems.} 
Subsequently, we assess the performance of TuRBO-1-LogEI-qREI in high-dimensional problems where restarts are infrequent. In the rover trajectory planning problem, as shown in Fig. \ref{fig:high-prob}, one or two restarts barely occurred, and with the performance improvements brought by qREI, both variants of TuRBO-1-LogEI-qREI achieved the best results after more than 1000 evaluations. For the HPA101-2 and MOPTA08 problems, almost no restarts occurred, and even when a restart did happen, the number of evaluations reached 2000 in the early stages of local search in the new trust region. Therefore, the effect of qREI was limited to the initial trust region selection, resulting in nearly identical performance between TuRBO-1-LogEI-qREI(restart) and TuRBO-1-LogEI. 

In the HPA101-2 problem, however, the introduction of qREI increased the probability of selecting a reliable initial trust region, and in the later stages of the optimization process, TuRBO-1-LogEI-qREI outperformed TuRBO-1-LogEI.
It should be noted that by increasing the batch size and the number of evaluations, the performance improvement effect of qREI during restarts has been confirmed, as detailed in Appendix \ref{apx:high-budget}.
The MOPTA problem exhibits notably smaller variability across trials than other problems and converges to solutions of similar quality regardless of the chosen trust region, thus resulting in a very limited effect of qREI. Additionally, because computation was terminated when TuRBO-1-LogEI-qREI caught up with TuRBO-1-LogEI, the effect of qREI is evident only in the reduction of the maximum values in the violin plot for TuRBO-1-LogEI-qREI compared to TuRBO-1-LogEI.

The approach using qREI is considered effective even for high-dimensional problems. However, in high-dimensional problems, the number of evaluations required before restarting increases, leading to fewer restarts, which is a limitation of this method. To address this, similar to the dimensionality management in Bounce, it may be effective to limit the number of function evaluations in each trust region and to shrink the trust region in a planned manner.

DSP and SAASBO showed good performance in the early stages of exploration in these high-dimensional problems although they had modest performance in medium-dimensional problems. 
While Bounce demonstrated excellent performance on HPA101-2, it did not perform as well on other problems. 
Since many of the methods in prior studies make assumptions about the problem characteristics, they may not always achieve sufficient performance depending on the nature of the real-world problem. Therefore, TuRBO-1-LogEI-qREI, which consistently shows stable and superior performance regardless of individual problem characteristics, can be considered a powerful method for medium to high-dimensional black-box optimization where the problem characteristics are unknown in advance.

\section{Conclusions}
In this paper, we introduced the use of regional expected improvement (REI) and its variant qREI within the TuRBO framework to enhance high-dimensional Bayesian optimization. We theoretically demonstrated the effectiveness of the region-averaged acquisition function, including qREI. Our experiments showed that when the trust region was restarted multiple times, the proposed method outperformed existing methods due to the optimal trust region selection by qREI. They also demonstrated robustness across different problem settings, in contrast to other methods that are more dependent on specific problem characteristics. These findings highlight qREI coupled with trust-region-based Bayesian optimization as a powerful and versatile tool for medium to high-dimensional black-box optimization.

\bibliography{mybib}

\clearpage
\appendix

\section*{Appendices}
\section{Proof}
\newcommand{\bfx}{\mathbf{x}}
\label{apx:app-proof}


We recall the following well-known results (\Cref{prop:rkhs-norm-l2,prop:rkhs-norm-restr}) on RKHS. 
For example, we refer to \cite[Chapters 10.2 and 10.7]{wendland2004scattered}.
\begin{proposition}
    \label{prop:rkhs-norm-l2}
    For $\extf{f} \in \cH_\Phi(\RR^D)$, the RKHS norm $\|\extf{f}\|_{\cH_\Phi(\RR^D)}$ is given as 
    \begin{equation}
        \label{eq:rkhs-norm-l2}
        \|\extf{f}\|_{\cH_\Phi(\RR^d)}^2
        = \|\cF \extf{f}/\sqrt{\cF \Phi}\|_{L_2(\RR^D)}^2 = 
        \int_{\RR^D} \frac{|\cF \extf{f}(\bfx)|^2}{\cF \Phi (\bfx)} d\bfx,
    \end{equation}
    where $\cF$ denotes the Fourier transform, i.e., for $f\in L_1(\RR^D)$,
    \begin{equation*}
        \cF f(\omega) = \int_{\bfx \in \RR^D} f(\omega) \exp(-i \bfx \cdot \omega)d\bfx.
    \end{equation*}
\end{proposition}

\begin{proposition}
    \label{prop:rkhs-norm-restr}
    For any $f \in \cH_{\Phi}(\Omega)$,
    there exists a natural extension $\extf{f} \in \cH_{\Phi}(\RR^D)$ of $f$ such that 
    $\|f\|_{\cH_\Phi(\Omega)} = \|\extf{f}\|_{\cH_{\Phi}(\RR^D)}$.
    Moreover, 
    for any $g \in \cH_\Phi(\RR^d)$ with $g|_{\Omega} = f \in \cH_\Phi(\Omega)$, 
    we have
    \begin{equation*}
        \|f\|_{\cH_\Phi(\Omega)} \le \|g\|_{\cH_\Phi(\RR^D)}.
    \end{equation*}
\end{proposition}

Next, we rewrite the operator $\cS$ using a convolution.
The following proposition follows from the definition of the operator $\cS$.
\begin{proposition}
    \label{prop:s-op-conv}
    Let $\chi_0: \RR^D \rightarrow \RR$ be the normalized indicator function of a trust region centered at the origin, i.e.,
    \begin{math}
        \chi_0 = (\prod_{i=1}^{D}l_i)^{-1} \indc_{\cT_0},
    \end{math}
    where $\cT_0 = \left(\prod_{i=1}^{D}[-l_i/2, l_i/2]\right)$.
    Then, by definition, the region-averaging operator $\cS : \cH_{\rbf}(\Omega) \rightarrow \cH_{\rbf}(\Omega)$ is given as 
    \begin{equation*}
    \cS f = (\extf{f} \ast \chi_0)|_{\Omega},
    \end{equation*}
    where $(f \ast g)(\bfx) = \int_{\bfx' \in \RR^d} f(\bfx')g(\bfx-\bfx')d\bfx$ 
    denotes the convolution of functions $f, g$. 
\end{proposition}
Now, we are ready to prove our main theoretical result.
\begin{proof}[Proof of \Cref{thm:reduced-upper-bound}]
    By definition of $U_T(f)$,
    it is sufficient to prove $\| \cS f\|_{\cH_\Phi(\Omega)} \le \|f \|_{\cH_\Phi(\Omega)}$.
    We note that $\cF \chi_0(\bfx)$ is given as 
    \begin{math}
        \prod_{i=1}^{D} \frac{\sin \left(2^{-1}\pi \lambda_i \bfx_i\right)}{2^{-1}\pi \lambda_i \bfx_i}.
    \end{math}
    It follows that 
    \begin{equation}
        \label{eq:cf-chi0-le1}
        |\cF \chi_0 (\bfx)| \le 1, \quad \forall \bfx \in \RR^D.
    \end{equation}
    Thus, we have the following
    \begin{align}
        \|\cS f\|_{\cH_\Phi(\Omega)}^2 &= 
        \| (\extf{f} \ast \chi_0)|_{\Omega}\|_{\cH_\Phi(\Omega)}^2 \label{eq:pr-sop-def}\\
        &\le \| \extf{f} \ast \chi_0\|_{\cH_\Phi(\RR^D)}^2 \label{eq:pr-restr}\\
        &=
        \int_{\RR^D} \frac{|\cF \extf{f}(\bfx)|^2|\cF \chi_0(\bfx)|^2}{\cF \Phi (\bfx)} d\bfx \label{eq:pr-l2-norm-conv}\\
        &\le 
        \int_{\RR^D} \frac{|\cF \extf{f}(\bfx)|^2}{\cF \Phi (\bfx)} d\bfx
        = \|f \|_{\cH_\Phi(\Omega)} ^2 \label{eq:pr-dirichlet}
    \end{align}
    Here, \eqref{eq:pr-sop-def} follows from \Cref{eq:pr-sop-def},
    \eqref{eq:pr-restr} follows from \Cref{prop:rkhs-norm-restr},
    \eqref{eq:pr-l2-norm-conv} follows from \Cref{prop:rkhs-norm-l2},
    the inequality in \eqref{eq:pr-dirichlet} follows from inequality \eqref{eq:cf-chi0-le1} 
    noting that $\cF \Phi(x)$ is non-negative by Bochner's theorem (c.f. \cite[Theorem 6.6]{wendland2004scattered}).
    This completes the proof.
\end{proof}

\subsection{Further Remarks on the Theoretical Result}
\paragraph{Gaussian Process and RKHS.}
We consider a GP model only for algorithm design, and assume an RKHS model in the theoretical analysis.
It is well-known that GP and RKHS are closely related and its connection is well-studied \cite{kanagawa2018gaussian}.
To discuss the connection, for simplicity, 
we assume that $k=k_\nu$ is a Mat\`ern kernel with a smoothness parameter $\nu > 0$, i.e., 
$k_\nu(\bfx, \bfx') = \frac{1}{2^{\nu - 1}\Gamma(\nu)} \left(\frac{\sqrt{2 \nu} \| \bfx - \bfx'\|}{h} \right)^{\nu}
K_\nu\left(\frac{\sqrt{2 \nu} \| \bfx - \bfx'\|}{h} \right)$, where $K_\nu$ is the modified Bessel function of the second kind 
and $h > 0$ is a length scale.
Then, if $f \sim \mathcal{GP}(0, k_\nu)$, for any $0 < \nu' < \nu$,
there is a version of $f$ that belongs to the RKHS corresponding to $k_\nu'$ with probability $1$ 
(see \citet[Corollary 4.15 and Remark 4.15]{kanagawa2018gaussian}).
Informally, this implies that 
for a given GP $f \sim \mathcal{GP}(0, k_\nu)$, if we take a less smooth kernel $k_{\nu'}$, a sample path of $f$ belongs to 
the RKHS $\cH_{k_{\nu'}}(\Omega)$ corresponding to $k_{\nu'}$.
Thus, by replacing the kernel $k$ by a less smooth kernel $k'$ in the theoretical analysis, 
we can improve the consistency between the algorithm design and theoretical analysis.

\paragraph{Definition of the Problem Complexity.}
In the theoretical analysis, we consider the regret upper bound $U_N(f)$ to discuss the problem complexity.
However, one could argue that lower bounds (or minimax optimal regret) would be more suitable notion to capture the problem complexity.
By the proof of \Cref{thm:reduced-upper-bound}, we can deduce a similar statement for lower bounds as follows.
For $B > 0$, we define a set $\cH_{\Phi, B}(\Omega)$ of objective functions by 
$\{f \in \cH_{\Phi}(\Omega): \|f\|_{\cH_{\Phi}(\Omega)} \le B\}$.
We assume that $f \in \cH_{\Phi, B}(\Omega)$ and $\cS f \in \cH_{\Phi, B'}(\Omega)$ with $0 < B' \le B$.
For instance, one can take $B= \|f \|_{\cH_\Phi(\Omega)}$ and $B' = \|\cS f \|_{\cH_\Phi(\Omega)}$. 
Then, by the proof of \cref{thm:reduced-upper-bound}, we have $B' \le B$.
By \cite[Theorem 2]{scarlett2017lower}, regret lower bound for the function space $\cH_{\Phi, B'}(\Omega)$
is less than or equal to that of $\cH_{\Phi, B}(\Omega)$, i.e.,
this result also indicates that the problem complexity can be reduced if we take the operator $\cS$.

\section{Parameter Setting}
\label{apx:setting}

\paragraph{Common Setting.}
In the implementations of TuRBO, GP-LogEI, and SAASBO using Botorch, all methods employ GP models with a Mat\'ern-5/2 kernel and automatic relevance determination (ARD). Sobol sequences are used for the initial sample generation. The random seed is set using the run number from 1 to 11. These settings for DSP and Bounce are configured according to their respective source codes.

\paragraph{TuRBO Variants.}
Another parameter related to the trust region is the maximum value of $l$ when the trust region expands, which is given by $l_{max}=1.6$. When using LogEI as the acquisition function for local search, the maximization of LogEI is performed using the L-BFGS-B method with 512 initial samples and 10 restarts, following the default settings in the Botorch implementation. When using TS, 2,000 to 5,000 samples (depending on the dimension $D$) are generated within the trust region by replacing some variables of the trust region center $\mathbf{c}$ with random numbers generated by Sobol sequences, and the sample where the TS function takes its minimum value is used for the actual function evaluation. The maximization of qREI is similar to that of LogEI, where Sobol sequences are also used to generate $N_x=128$ samples $\hat{\mathbf{x}}_j$ within the trust region candidate. 

\paragraph{GP-LogEI.}
LogEI is maximized using the L-BFGS-B method with 512 initial samples and 10 restarts, in conjunction with TuRBO.

\paragraph{SAASBO.}
In the original paper, the acquisition function used is EI; however, we have replaced it with LogEI. The maximization of LogEI is performed using the L-BFGS-B method with 512 initial samples and 10 restarts, in conjunction with TuRBO.

\paragraph{Bounce.}
For all problems, the initial search is conducted in the 5-dimensional space, and the maximum number of samples used before reaching the full-dimensional space of the problem is set to 500.

\paragraph{DSP.}
Following the source code provided in the original paper, we employ the Gaussian (RBF) kernel and the qLogNEI acquisition function, which is a noisy variant of the qLogEI acquisition function.

\section{Statistical Testing}
\label{apx:testing}
\begin{figure}[tb]
    \centering
    \includegraphics[width=0.9\columnwidth]{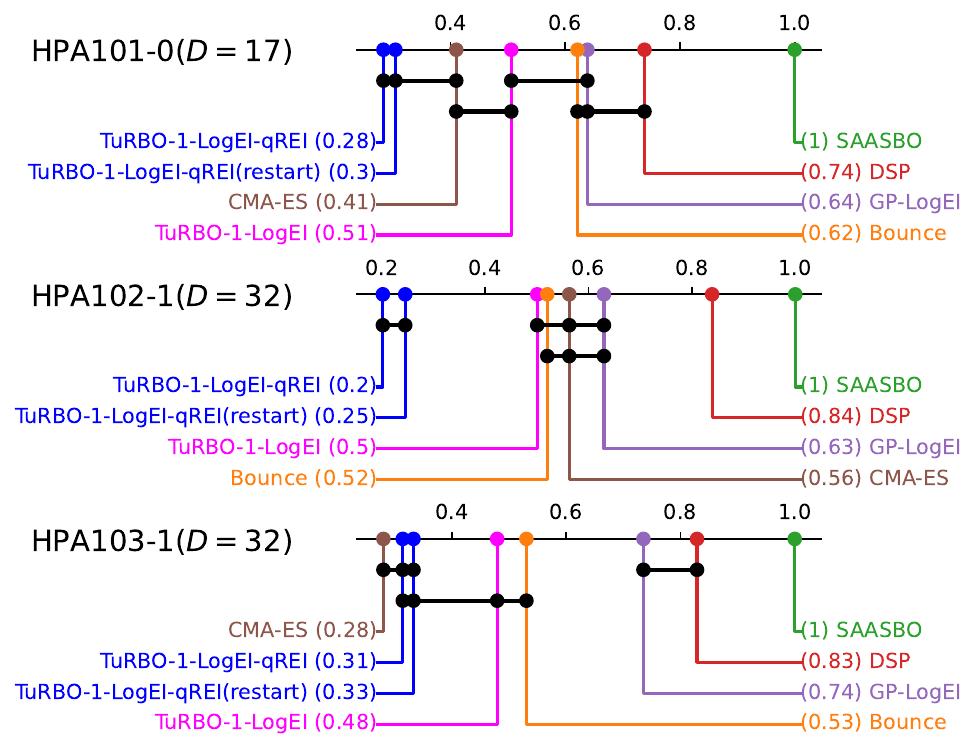}
    \caption{Critical difference diagram for the medium-dimensional problems}
    \label{fig:test-mid}
\vspace{4mm}
    \includegraphics[width=0.9\columnwidth]{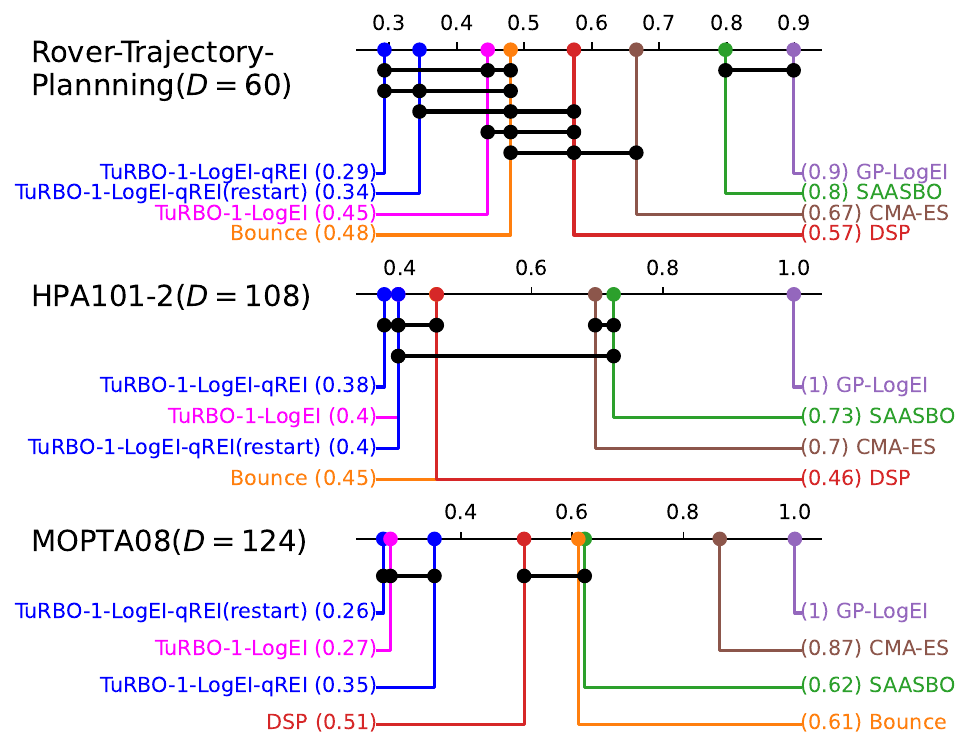}
    \caption{Critical difference diagram for the high-dimensional problems}
    \label{fig:test-high}
\end{figure}

The results of the Wilcoxon signed-rank test and rank-sum test at a significance level of 5\% for the final objective function values obtained by each method are presented in Figs. \ref{fig:test-mid} and \ref{fig:test-high}. No correction for multiple comparisons was applied. In comparisons among the TuRBO variants, GP-LogEI, and SAASBO, the initial samples for each run are identical, so the paired Wilcoxon signed-rank test was used. For other tests where there is no such correspondence, the Wilcoxon rank-sum test was applied.
In medium-dimensional problems, TuRBO-1-LogEI-qREI variants showed significant differences compared to TuRBO-1-LogEI in the HPA101-0 and 102-1 problems. In the high-dimensional problem, a significant difference was observed between TuRBO-1-LogEI-qREI(restart) and TuRBO-1-LogEI in the rover trajectory planning problem.

\section{Effectiveness of Region-Averaged Acquisition Functions}
\label{apx:rei-vs-ei}
\begin{figure*}[ht]
\centering
\includegraphics[width=0.9\textwidth]{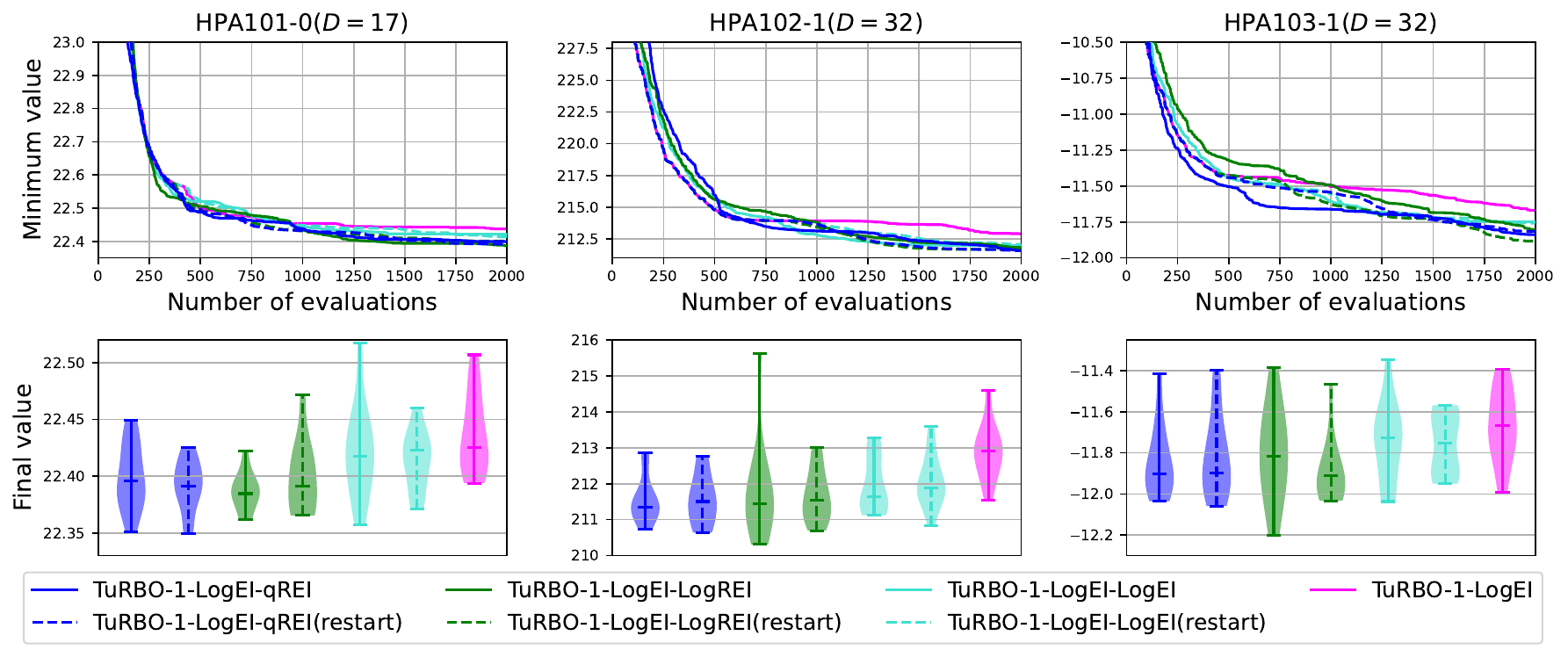} 
\caption{Comparison among qREI, LogREI, and LogEI.}
\label{fig:rei-vs-ei}
\vspace{4mm}
\centering
\includegraphics[width=0.9\textwidth]{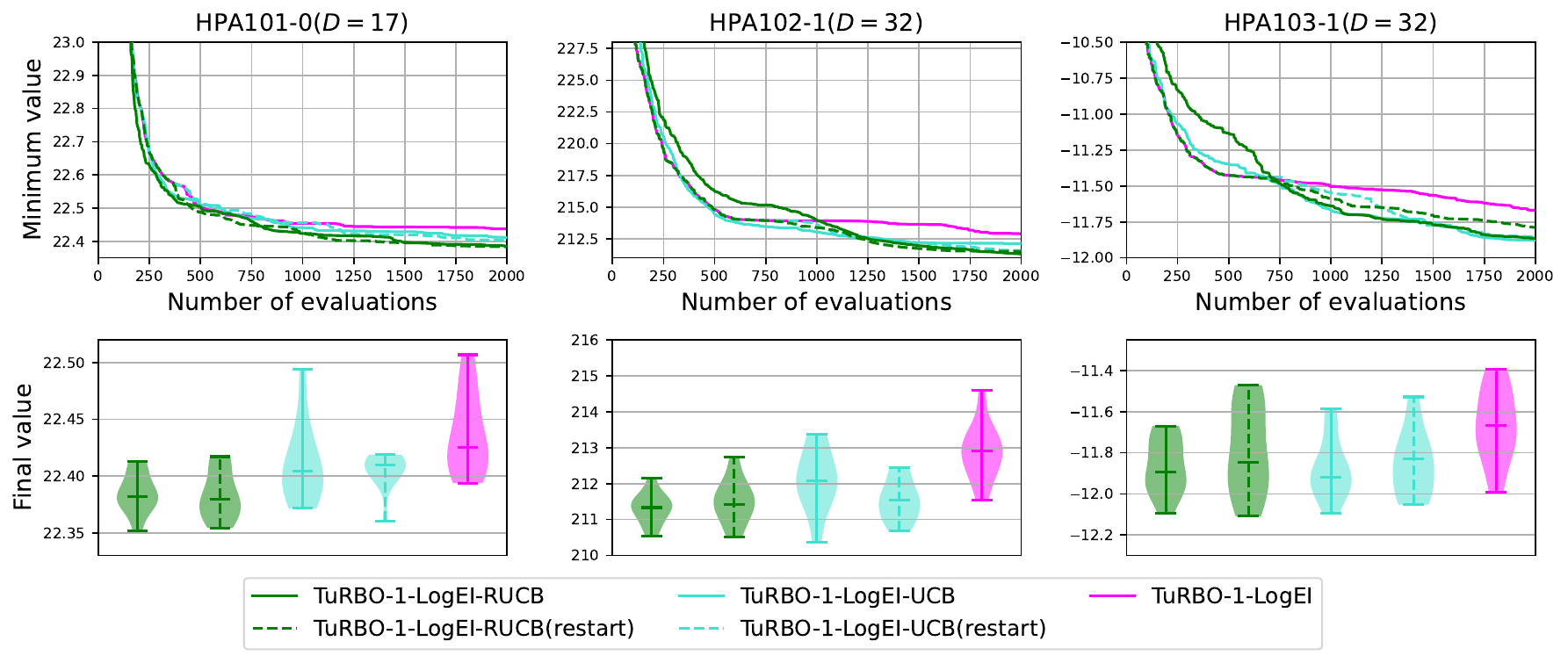} 
\caption{Comparison between RUCB and UCB.}
\label{fig:rucb-vs-ucb}
\vspace{4mm}
\centering
\includegraphics[width=0.9\textwidth]{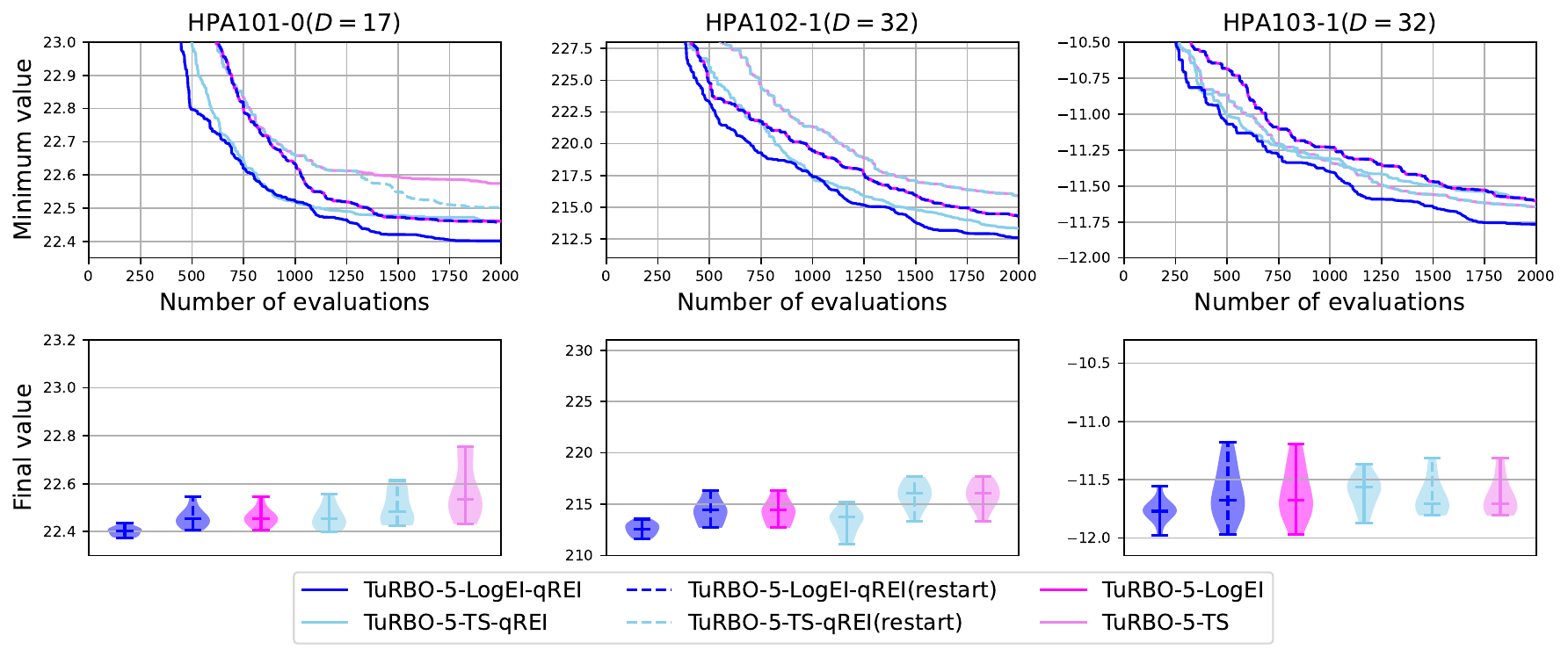} 
\caption{History of TuRBO-5 performances for the medium-dimensional problems.}
\label{fig:turbo5}
\end{figure*}

We experimentally demonstrate that the region-averaged acquisition functions improve optimization performance compared to the trust region selection based on the conventional acquisition functions. As a baseline for comparison, we use a method that determines the trust region center by maximizing LogEI, instead of qREI. This method is denoted as TuRBO-1-LogEI-LogEI. We also use TuRBO-1-LogEI-LogREI, where region-averaged LogEI is employed for trust region selection, for fair comparison. Fig. \ref{fig:rei-vs-ei} show the comparison results for the medium-dimensional problems.

Although using LogEI for trust region selection improves performance compared to TuRBO-1-LogEI, the violin plots indicate that trust region selection using LogREI and qREI is superior in terms of final objective function values. As shown in Fig. \ref{fig:rucb-vs-ucb}, a similar result is observed when comparing TuRBO-1-LogEI-UCB, which employs a standard UCB acquisition function for trust region selection, with TuRBO-1-LogEI-RUCB, which utilizes a region-averaged UCB (RUCB). This experimental evidence supports the theoretical analysis of the effectiveness of region-averaged acquisition functions and confirms the validity of using REI for trust region selection.

\section{qREI for TuRBO-$m$}
\label{apx:trubom}
When using qREI with TuRBO-$m$ ($m>1$), the initial trust region is determined simultaneously by maximizing qREI with $q=m$. On the other hand, since restarts are performed individually for each trust region, qREI with $q=1$ is used during restarts. Fig. \ref{fig:turbo5} shows the results of applying TuRBO-5 ($m=5$) to the medium-dimensional problems. The batch size is also 1 in TuRBO-5. TuRBO-5-LogEI/TS-qREI demonstrates superior performance compared to TuRBO-5-LogEI/TS in most cases. In TuRBO-5, the distribution of sample points across the five trust regions reduces the frequency of restarts compared to TuRBO-1, and in most cases, the performance of TuRBO-5-LogEI/TS-qREI(restart) is generally consistent with that of TuRBO-5-LogEI/TS. In cases where the number of evaluations is up to 2,000, the performance of TuRBO-1 generally surpasses that of TuRBO-5.

\section{qREI for TuRBO-1 with Thompson Sampling}
\label{apx:trubo-ts}
\begin{figure*}[ht]
\centering
\includegraphics[width=0.9\textwidth]{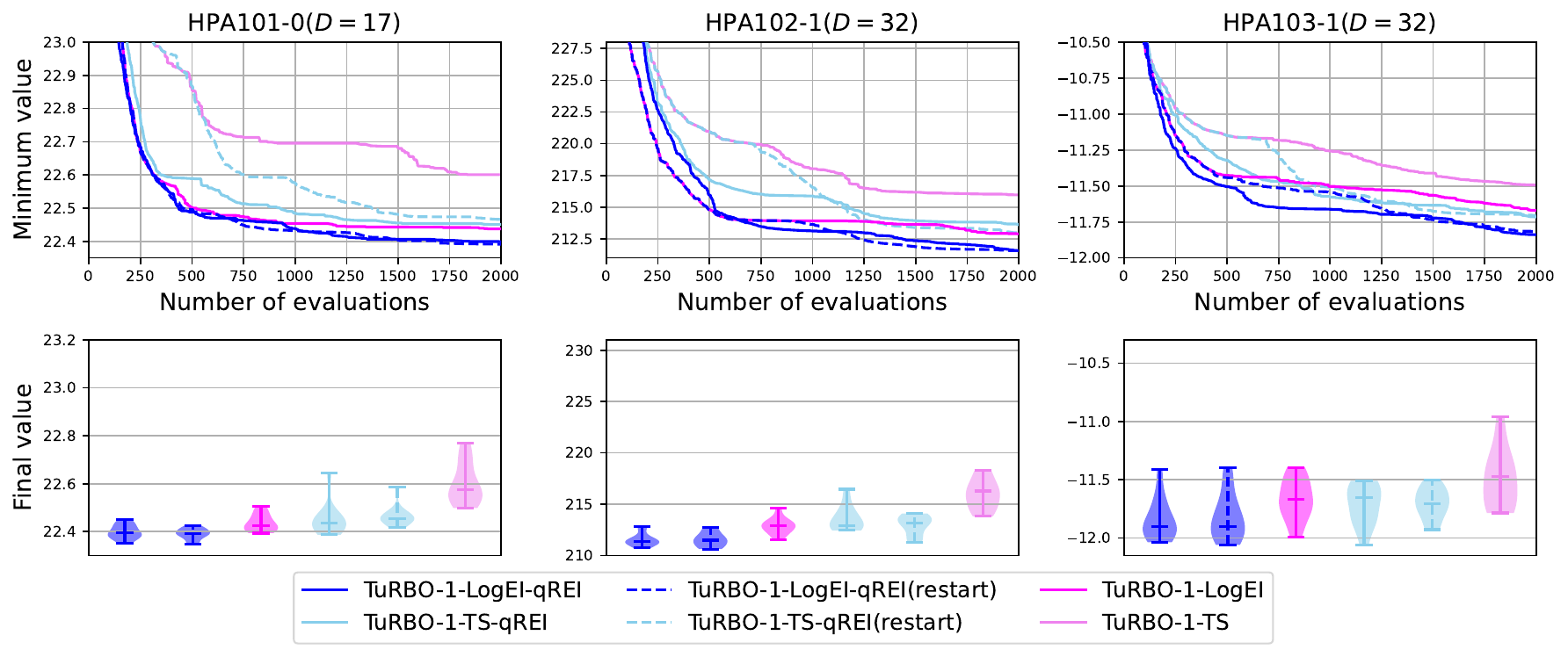} 
\caption{Performance comparison between TuRBO with LogEI and TS for the medium-dimensional problems.}
\label{fig:ts-mid}
\vspace{4mm}
\centering
\includegraphics[width=0.9\textwidth]{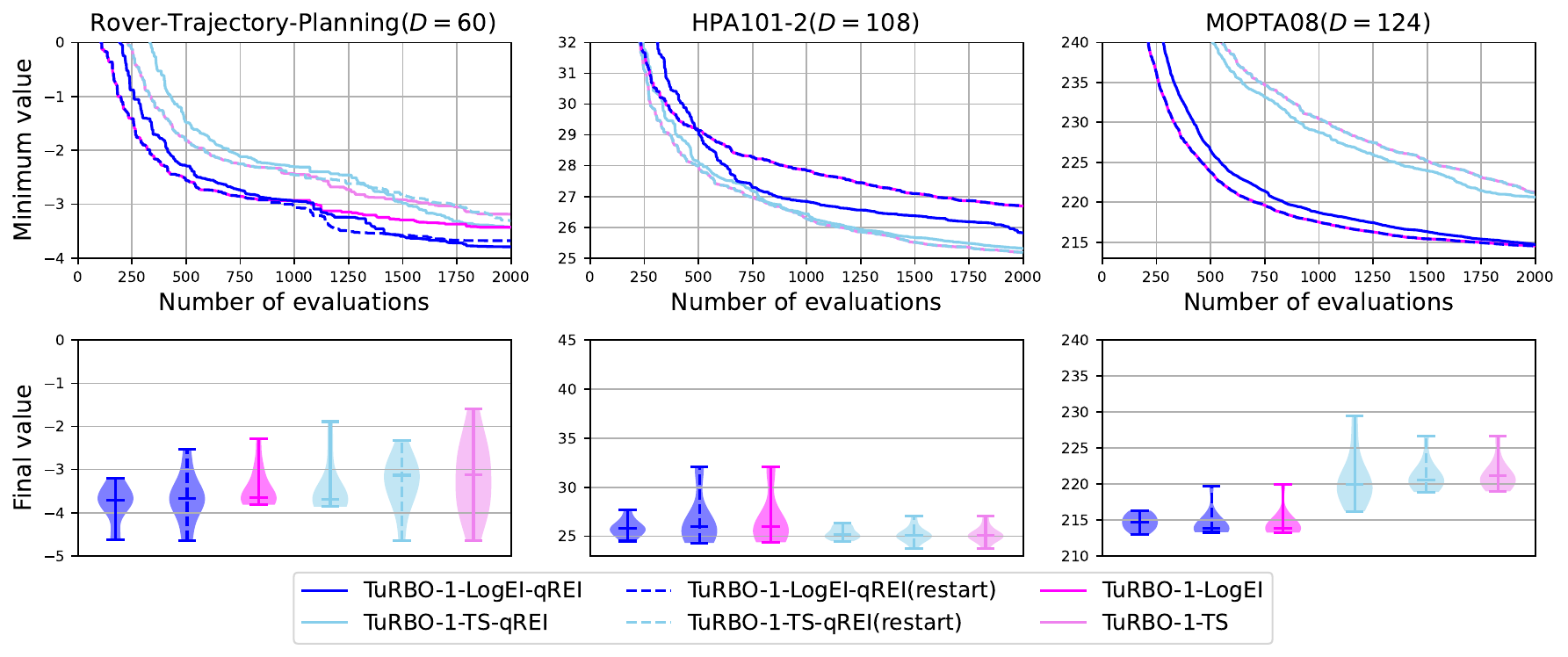} 
\caption{Performance comparison between TuRBO with LogEI and TS for the high-dimensional problems.}
\label{fig:ts-high}
\vspace{4mm}
\centering
\includegraphics[width=0.6\textwidth]{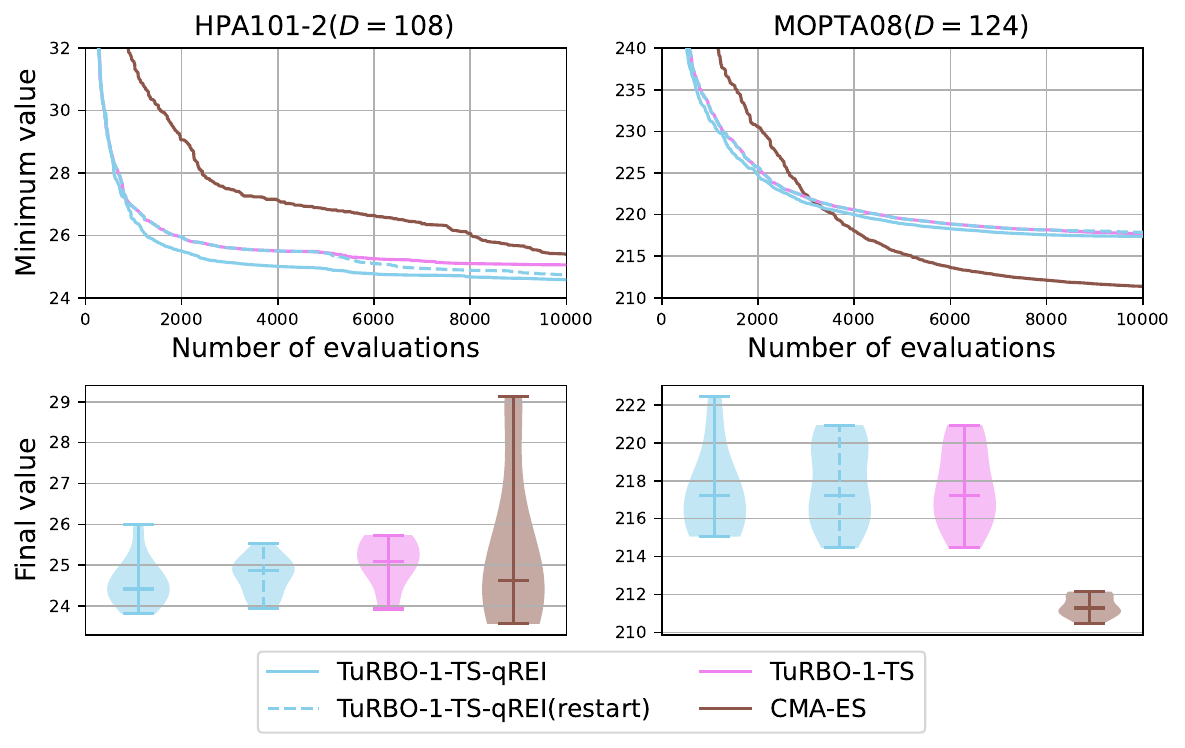} 
\caption{History of objective function values for the HPA101-1 problem up to 10,000 function evaluations.}
\label{fig:high10000}
\end{figure*}

We compare the results of using TS as the acquisition function in the local search of TuRBO with those obtained using LogEI. As shown in Figs. \ref{fig:ts-mid} and \ref{fig:ts-high}, performance improvements due to the introduction of qREI can be observed even when using TS. However, regardless of the presence of qREI, TuRBO with LogEI generally shows better performance, except in the HPA101-2 problem.

\section{High-Budget Performance}
\label{apx:high-budget}
In high-dimensional problems, due to the minimal occurrence of trust region restarts in TuRBO within 2,000 function evaluations, we conducted experiments with 10,000 evaluations on the HPA101-2 and MOPTA08 problems. To reduce computational costs, the batch size was set to 5, and only TuRBO with TS was used, as sample additions using qLogEI are computationally expensive.

To further reduce the cost of constructing the GP model as the number of samples increases, a subset of $N_{\text{GP}}$ representative points was extracted from the previously evaluated sample points and used in the GP for both the local search and the trust region selection with qREI. To extract these representative points, the following log regret $r_i$ was calculated for the dataset used ($\mathcal{D} = \mathcal{D}_{t}$ for trust region selection or $\mathcal{D} = \mathcal{D}^{TR}$ for local search):
\begin{align}
    \label{eq:regret}
    & r_i = \frac{\log (\bar{f}_i + \bar{f}_{\text{min}}) - \log \bar{f}_{\text{min}}}{\log (1+\bar{f}_{\text{min}}) - \log \bar{f}_{\text{min}}}, \\
    & \bar{f}_{\text{min}} = \min_{\bar{f}_i \in \mathcal{D}} (\bar{f}_i | \bar{f}_i > 0), \quad
      \bar{f}_i = \frac{f_i - f_{\text{min}}}{f_{\text{max}} - f_{\text{min}}}
\end{align}
where, $f_{\text{min}}$ and $f_{\text{max}}$ represent the minimum and maximum objective function values within $\mathcal{D}$. Representative points are then greedily selected in the ($D+1$) dimensional space formed by $r_i$ and the normalized design variables $\mathbf{x}_i$. Specifically, the sample with $r_i = \bar{f}_i = 0$ is selected first as a representative point, and subsequently, the unselected sample point with the max-min distance to all previously selected representative points is added to the set until $N_{\text{GP}}$ representative points are selected. Here, we use $N_{\text{GP}}=500$.

The optimization results using these methods for reducing computational costs are shown in Fig. \ref{fig:high10000}. In the HPA101-2 problem, by increasing the number of evaluations, the benefits of using qREI for the initial trust region selection and restarts become evident, leading to improved objective function values compared to cases where qREI is not used. 
In contrast, for the MOPTA08 problem, restarts occurred around the 8,000th evaluation, so within the 10,000 evaluations considered, we could not confirm their effectiveness. In this problem, increasing the number of evaluations leads to improved performance of CMA-ES, suggesting that exploitation within a single trust region is more important than exploration.

\section{Low-Budget Performance}
\label{apx:low-budget}
\begin{figure*}[ht]
\centering
\includegraphics[width=0.9\textwidth]{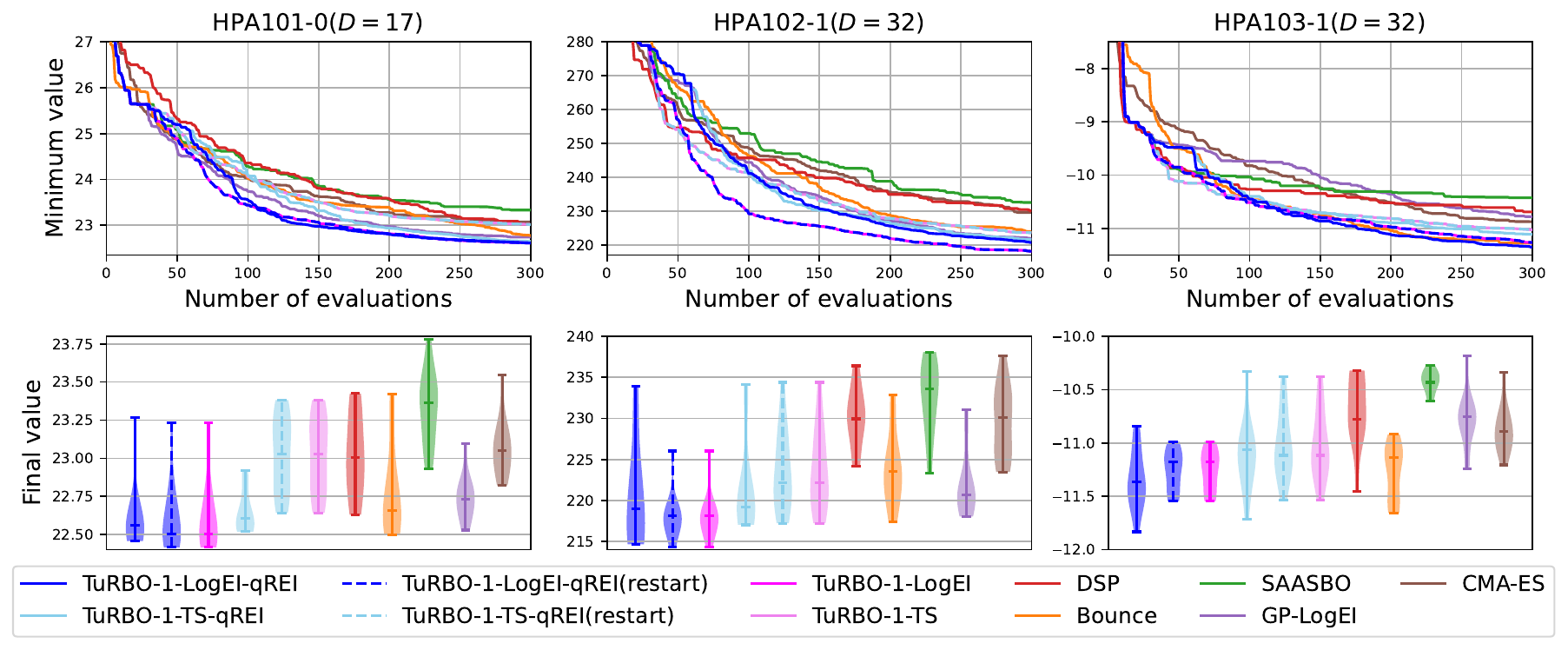} 
\caption{History of objective function values for the medium-dimensional problems within 300 function evaluations.}
\label{fig:mid300}
\vspace{4mm}
\centering
\includegraphics[width=0.9\textwidth]{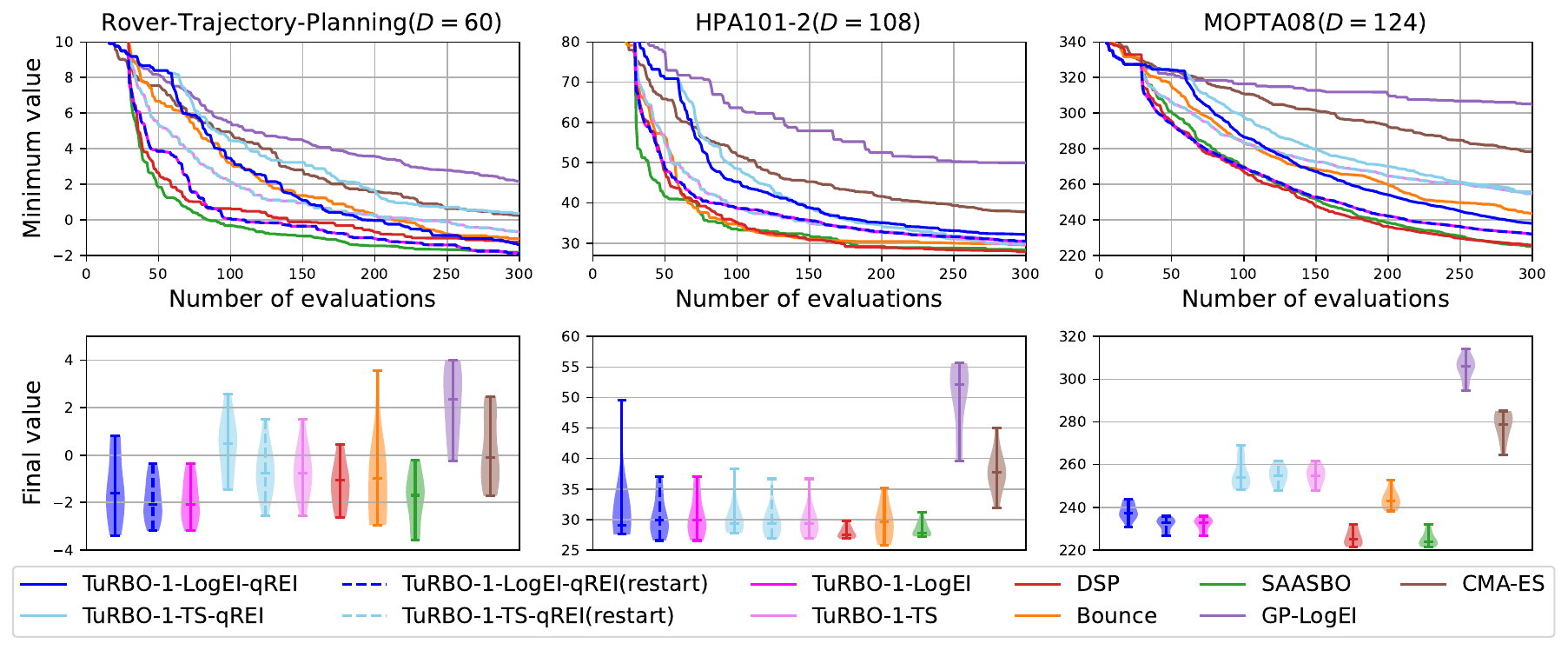} 
\caption{History of objective function values for the high-dimensional problems within 300 function evaluations.}
\label{fig:high300}
\vspace{4mm}
\centering
\includegraphics[width=0.9\textwidth]{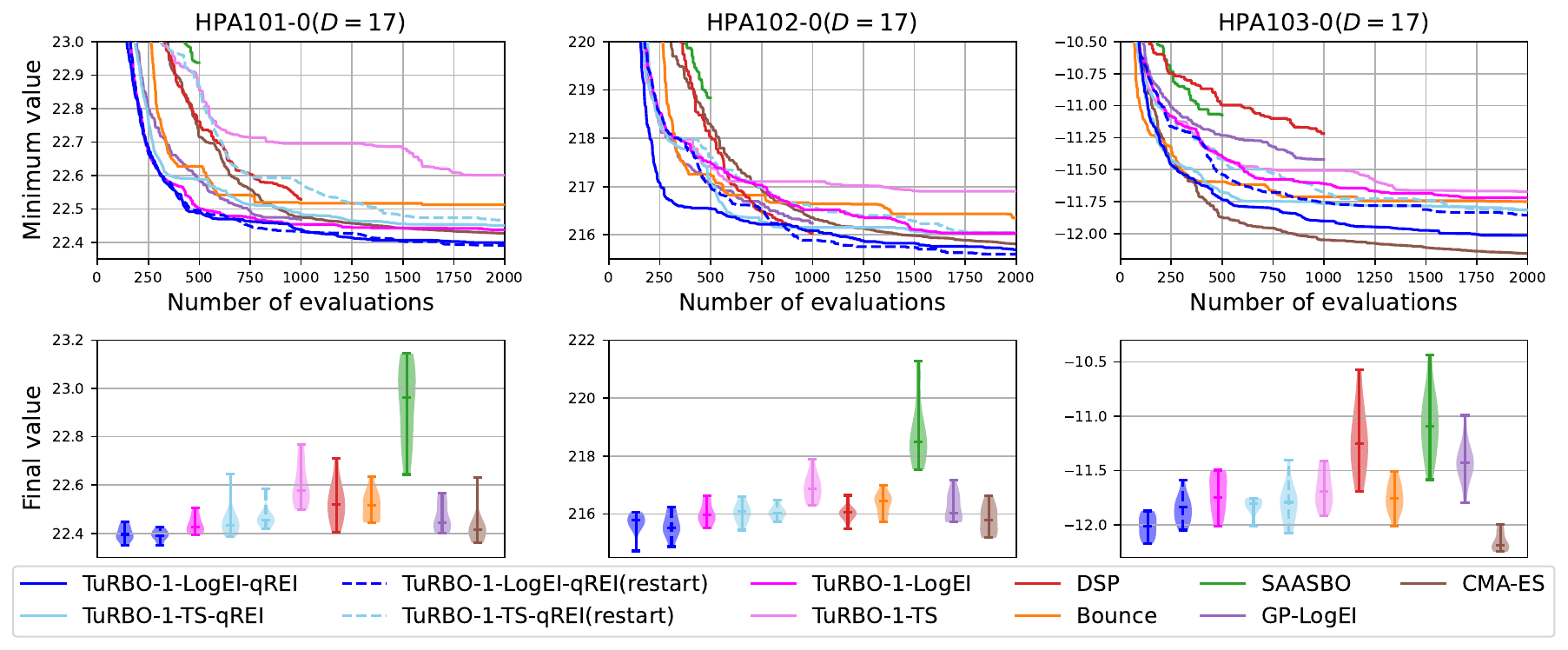} 
\caption{History of objective function values for the 17-dimensional HPA problems.}
\label{fig:hpa0}
\end{figure*}

The performance of each method with up to 300 function evaluations is shown in Figs. \ref{fig:mid300} and \ref{fig:high300}. Within this range, no restarts occur in TuRBO, so the performance of TuRBO-1-LogEI/TS-qREI(restart) is identical to that of TuRBO-1-LogEI/TS. Using qREI for the initial trust region selection reduces the number of sample points available for local search by 30, leading to a performance decrease in the HPA102-1 and high-dimensional problems. Notably, TuRBO-1-LogEI demonstrates the best performance in medium-dimensional problems and also achieves performance comparable to DSP and SAASBO in high-dimensional problems with relatively low computational cost, indicating that TuRBO is a promising method across a wide range of problems.

\section{Other HPA Problems}
\label{apx:other-problem}

The results of applying each method to the six HPA problems are shown in Figs. \ref{fig:hpa0} and \ref{fig:hpa1}. Two variants of TuRBO-1-LogEI/TS-qREI improve the performance of TuRBO-1-LogEI/TS across all problems.

\section{Computational Resources and Time}
\label{apx:time}
\begin{figure*}[ht]
\centering
\includegraphics[width=0.9\textwidth]{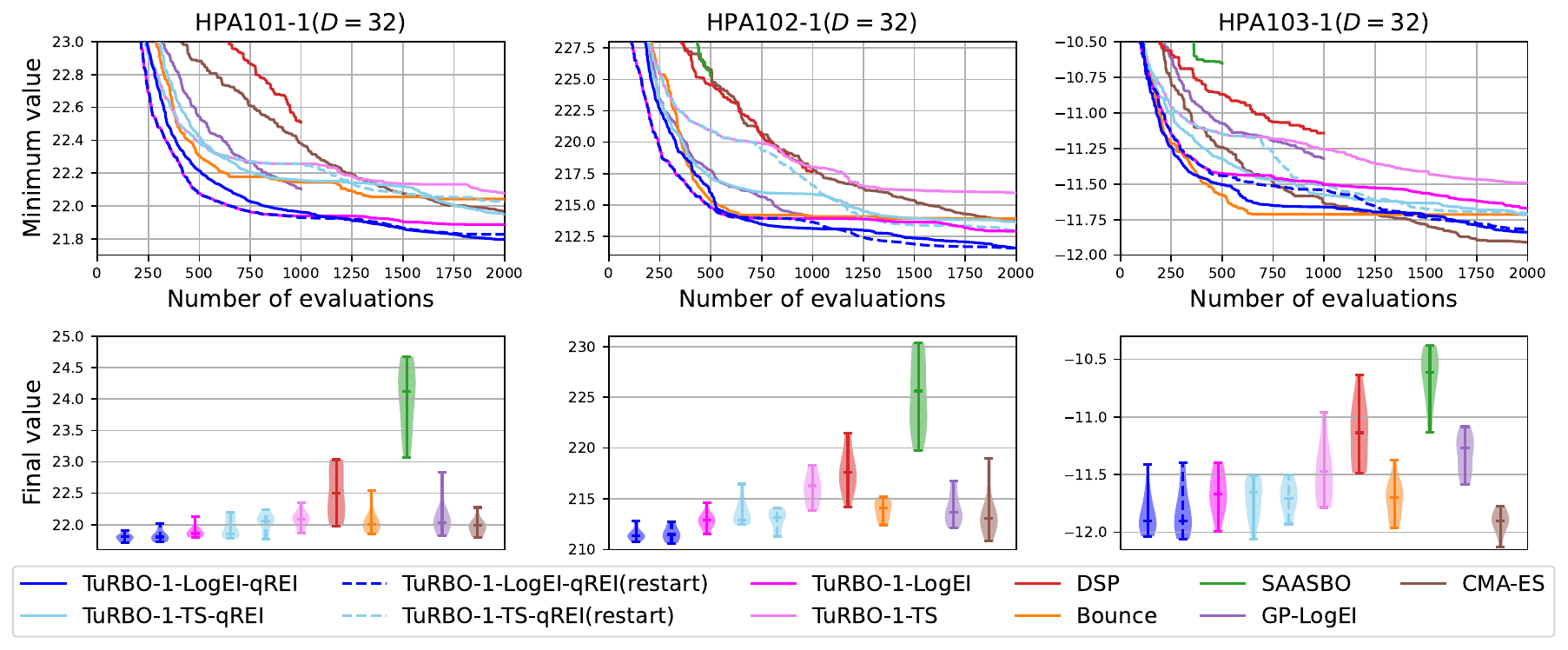} 
\caption{History of objective function values for the 32-dimensional HPA problems.}
\label{fig:hpa1}
\vspace{4mm}
\centering
\includegraphics[width=0.9\textwidth]{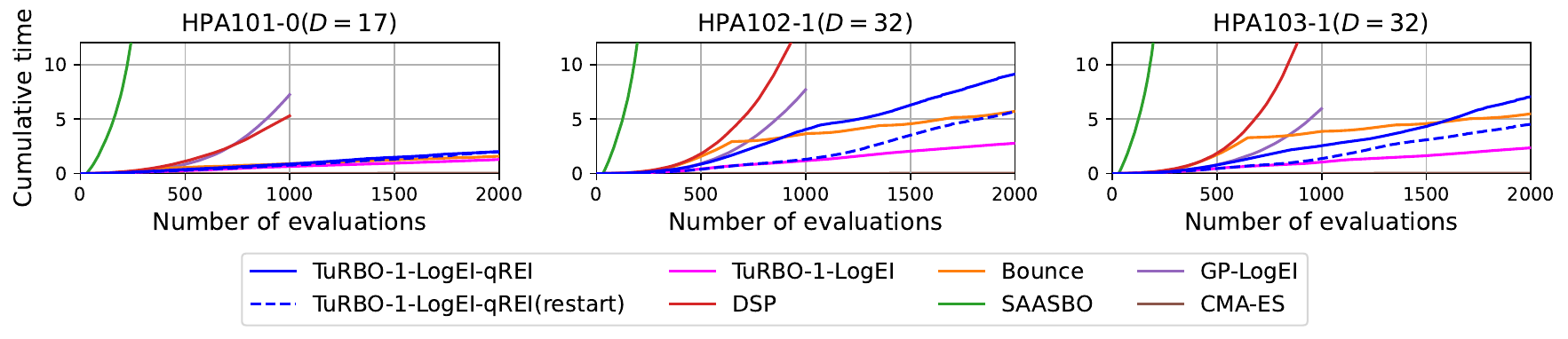}
\caption{Computaitonal time for medium-dimensional problems.}
\label{fig:mid-time}
\vspace{4mm}
\centering
\includegraphics[width=0.9\textwidth]{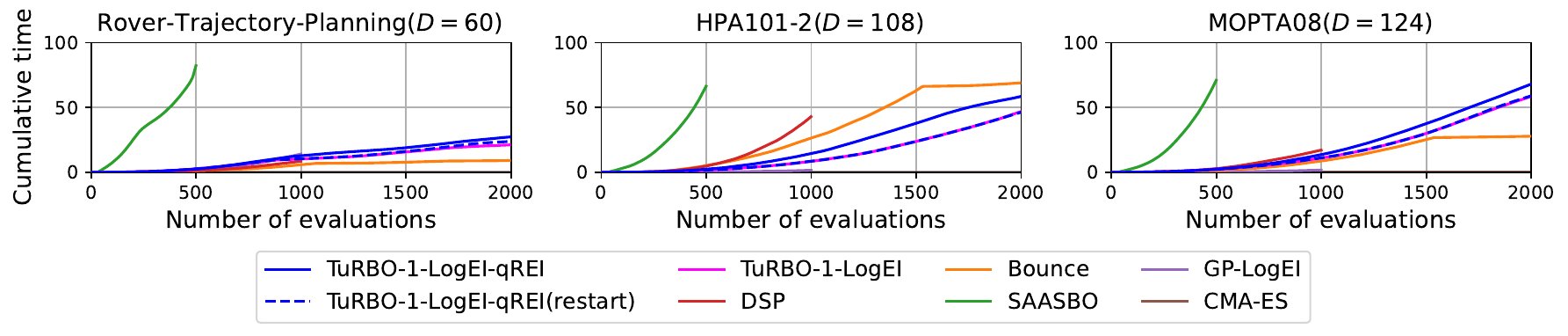}
\caption{Computaitonal time for high-dimensional problems.}
\label{fig:high-time}
\end{figure*}

The numerical experiments were conducted using Intel Xeon Gold 6148 2.4 GHz processors, with 32GB of memory. Figs. \ref{fig:mid-time} and \ref{fig:high-time} show the mean cumulative computation time over 11 runs as a function of the increasing number of sample points for each method.

The computation time of TuRBO variants varies depending on the presence of qREI. In many problems, even before restarts, the computation time for TuRBO-1-LogEI-qREI notably exceeds that of TuRBO-1-LogEI. This suggests that the construction of the GP and/or the maximization of LogEI takes longer in the trust regions selected by qREI. 
The computation time required to determine the trust region using qREI ranges from 2 to 20 minutes, depending on the number of samples. Even for the problem with the most restarts, HPA101-0, the number of restarts is 4 or 5, while HP102-1 and HPA103-1 require 1 to 4 restarts. Thus, the impact on overall computation time is limited. 

The computation time for DSP varies considerably depending on the problem. While it increases rapidly in the HPA problems, it remains comparable to TuRBO in the Rover Trajectory and MOPTA08 problems, despite their high dimensionality. SAASBO requires approximately 70 hours for 500 function evaluations across all problems, making it the most computationally expensive method. In Bounce, the increase in computation time after restarts is more gradual than at the start. GP-LogEI shows longer computation times for the medium-dimensional problems, where it performs relatively well, but extremely short computation times for the high-dimensional problems, where its performance is poor. The computation time for CMA-ES is negligibly small in the figures.

\end{document}